\documentclass{article}

\usepackage{microtype}
\usepackage{graphicx}
\usepackage{booktabs} 

\usepackage{hyperref}

\usepackage{amsmath}
\usepackage{amsfonts}
\usepackage{amssymb}
\usepackage{subfig}
\usepackage{enumitem}
\usepackage{float}
\usepackage{bbm}
\usepackage{tikz}
\usetikzlibrary{positioning,angles,quotes}
\usepackage{fancyvrb}
\usepackage{caption}
\newcommand{\e}{\epsilon}
\newcommand{\y}{\gamma}
\newcommand{\al}{\alpha}
\newcommand{\s}{\sigma}
\newcommand{\ta}{\theta}
\newcommand{\w}{\omega}
\newcommand{\E}{\mathbb{E}}
\newcommand{\N}{\mathcal{N}}
\newcommand{\lb}{\left [}
\newcommand{\rb}{\right ]}
\newcommand{\lp}{\left (}
\newcommand{\rp}{\right )}
\newcommand{\B}{\mathcal{B}}

\DeclareMathOperator*{\argmin}{argmin}
\DeclareMathOperator*{\argmax}{argmax}

\DeclareMathOperator{\oneargmax}{argmax}

\usepackage[accepted]{icml2019}

\icmltitlerunning{Off-Policy Deep Reinforcement Learning without Exploration}

\begin{document}

\twocolumn[
\icmltitle{Off-Policy Deep Reinforcement Learning without Exploration}

\begin{icmlauthorlist}
\icmlauthor{Scott Fujimoto}{mcgill,mila}
\icmlauthor{David Meger}{mcgill,mila}
\icmlauthor{Doina Precup}{mcgill,mila}
\end{icmlauthorlist}

\icmlaffiliation{mcgill}{Department of Computer Science, McGill University, Montreal, Canada}
\icmlaffiliation{mila}{Mila Qu\'ebec AI Institute}

\icmlcorrespondingauthor{Scott Fujimoto}{scott.fujimoto@mail.mcgill.ca}

\icmlkeywords{Machine Learning, ICML, Deep Reinforcement Learning, Imitation Learning, Batch Reinforcement Learning, Off-Policy}

\vskip 0.3in
]




\printAffiliationsAndNotice{} 

\begin{abstract}
Many practical applications of reinforcement learning constrain agents to learn from a fixed batch of data which has already been gathered, without offering further possibility for data collection. In this paper, we demonstrate that due to errors introduced by extrapolation, standard off-policy deep reinforcement learning algorithms, such as DQN and DDPG, are incapable of learning without data correlated to the distribution under the current policy, making them ineffective for this fixed batch setting. We introduce a novel class of off-policy algorithms, batch-constrained reinforcement learning, which restricts the action space in order to force the agent towards behaving close to on-policy with respect to a subset of the given data. We present the first continuous control deep reinforcement learning algorithm which can learn effectively from arbitrary, fixed batch data, and empirically demonstrate the quality of its behavior in several tasks.


\end{abstract}

\section{Introduction}

\textit{Batch reinforcement learning}, the task of learning from a fixed dataset without further interactions with the environment, is a crucial requirement for scaling reinforcement learning to tasks where the data collection procedure is costly, risky, or time-consuming. 
Off-policy batch reinforcement learning has important implications for many practical applications. It is often preferable for data collection to be performed by some secondary controlled process, such as a human operator or a carefully monitored program. If assumptions on the quality of the behavioral policy can be made, imitation learning can be used to produce strong policies. However, most imitation learning algorithms are known to fail when exposed to suboptimal trajectories, or require further interactions with the environment to compensate \cite{hester2017deep, sun2018truncated, cheng2018fast}. 
On the other hand, batch reinforcement learning offers a mechanism for learning from a fixed dataset without restrictions on the quality of the data.

Most modern off-policy deep reinforcement learning algorithms fall into the category of \textit{growing batch learning}~\citep{lange2012batch}, in which data is collected and stored into an experience replay dataset~\citep{expreplay1992}, which is used to train the agent before further data collection occurs. However, we find that these ``off-policy'' algorithms can fail in the batch setting, becoming unsuccessful if the dataset is uncorrelated to the true distribution under the current policy. Our most surprising result shows that off-policy agents perform dramatically worse than the behavioral agent \textit{when trained with the same algorithm on the same dataset}.

This inability to learn truly off-policy is due to a fundamental problem with off-policy reinforcement learning we denote \textit{\mbox{extrapolation} \mbox{error}}, a phenomenon in which unseen state-action pairs are erroneously estimated to have unrealistic values. 
Extrapolation error can be attributed to a mismatch in the distribution of data induced by the policy and the distribution of data contained in the batch. As a result, it may be impossible to learn a value function for a policy which selects actions not contained in the batch. 


To overcome extrapolation error in off-policy learning, we introduce batch-constrained reinforcement learning, where agents are trained to maximize reward while minimizing the mismatch between the state-action visitation of the policy and the state-action pairs contained in the batch. 
Our deep reinforcement learning algorithm, Batch-Constrained deep Q-learning (BCQ), uses a state-conditioned generative model to produce only previously seen actions. This generative model is combined with a Q-network, to select the highest valued action which is similar to the data in the batch. 
Under mild assumptions, we prove this batch-constrained paradigm is necessary for unbiased value estimation from incomplete datasets for finite deterministic MDPs. 


Unlike any previous continuous control deep reinforcement learning algorithms, BCQ is able to learn successfully without interacting with the environment by considering extrapolation error. Our algorithm is evaluated on a series of batch reinforcement learning tasks in MuJoCo environments \cite{mujoco,OpenAIGym}, where extrapolation error is particularly problematic due to the high-dimensional continuous action space, which is impossible to sample exhaustively.
Our algorithm offers a unified view on imitation and off-policy learning, and is capable of learning from purely expert demonstrations, as well as from finite batches of suboptimal data, without further exploration. We remark that BCQ is only one way to approach batch-constrained reinforcement learning in a deep setting, and we hope that it will be serve as a foundation for future algorithms. 
To ensure reproducibility, we provide precise experimental and implementation details, and our code is made available (\url{https://github.com/sfujim/BCQ}).

\vspace{-2mm}
\section{Background}

In reinforcement learning, an agent interacts with its environment, typically assumed to be a Markov decision process (MDP) $(\mathcal{S}, \mathcal{A}, p_M, r, \y)$, with state space $\mathcal{S}$, action space $\mathcal{A}$, and transition dynamics $p_M(s'|s,a)$. 
At each discrete time step, the agent receives a reward $r(s,a,s') \in \mathbb{R}$ for performing action $a$ in state $s$ and arriving at the state $s'$. The goal of the agent is to maximize the expectation of the sum of discounted rewards, known as the return $R_t = \sum_{i={t+1}}^\infty \y^i r(s_i, a_i, s_{i+1})$, which weighs future rewards with respect to the discount factor $\y \in [0,1)$. 

The agent selects actions with respect to a policy $\pi: \mathcal{S} \rightarrow \mathcal{A}$, which exhibits a distribution $\mu^\pi(s)$ over the states $s \in \mathcal{S}$ visited by the policy. Each policy $\pi$ has a corresponding value function $Q^\pi(s,a)=\E_\pi[R_t|s,a]$, the expected return when following the policy after taking action $a$ in state $s$. 
For a given policy $\pi$, the value function can be computed through the Bellman operator $\mathcal{T}^\pi$:
\begin{equation}
\mathcal{T}^\pi Q(s,a) = \E_{s'} [r + \y Q(s',\pi(s'))].
\end{equation}
The Bellman operator is a contraction for $\y \in [0,1)$ with unique fixed point $Q^\pi(s,a)$~\citep{bertsekas1995dynamic}. $Q^*(s,a) = \max_\pi Q^\pi(s,a)$ is known as the optimal value function, which has a corresponding optimal policy obtained through greedy action choices. 
For large or continuous state and action spaces, the value can be approximated with neural networks, e.g. using the Deep Q-Network algorithm (DQN)~\citep{DQN}. In DQN, the value function $Q_\ta$ is updated using the target: 
\begin{equation} \label{eqn:DQN}
r + \y Q_{\ta'}(s',\pi(s')), \quad \pi(s') = \oneargmax_a Q_{\ta'}(s',a),
\end{equation}
Q-learning is an \textit{off-policy} algorithm~\citep{sutton1998reinforcement}, meaning the target can be computed without consideration of how the experience was generated. 
In principle, off-policy reinforcement learning algorithms are able to learn from data collected by any behavioral policy. 
Typically, the loss is minimized over mini-batches of tuples of the agent's past data, $(s,a,r,s') \in \B$, sampled from an experience replay dataset $\B$~\citep{expreplay1992}. For shorthand, we often write $s \in \B$ if there exists a transition tuple containing $s$ in the batch $\B$, and similarly for $(s,a)$ or $(s,a,s') \in \B$. 
In batch reinforcement learning, we assume $\B$ is fixed and no further interaction with the environment occurs. 
To further stabilize learning, a target network with frozen parameters $Q_{\ta'}$, is used in the learning target. The parameters of the target network $\ta'$ are updated to the current network parameters $\ta$ after a fixed number of time steps, or by averaging $\ta' \leftarrow \tau \ta + (1 - \tau) \ta'$ for some small $\tau$ \citep{DDPG}. 

In a continuous action space, the analytic maximum of Equation (\ref{eqn:DQN}) is intractable. In this case, actor-critic methods are commonly used, where action selection is performed through a separate policy network $\pi_\phi$, known as the actor, and updated with respect to a value estimate, known as the critic \citep{sutton1998reinforcement, konda2003onactor}. This policy can be updated following the deterministic policy gradient theorem \cite{DPG}:
\begin{equation}
\phi \leftarrow \oneargmax_\phi \E_{s \in \B} [Q_\ta (s,\pi_\phi(s))], 
\end{equation}
which corresponds to learning an approximation to the maximum of $Q_\ta$, by propagating the gradient through both $\pi_\phi$ and $Q_\ta$. When combined with off-policy deep Q-learning to learn $Q_\ta$, this algorithm is referred to as Deep Deterministic Policy Gradients (DDPG) \citep{DDPG}.

\vspace{-2mm}
\section{Extrapolation Error}


Extrapolation error is an error in off-policy value learning which is introduced by \textit{the mismatch between the dataset and true state-action visitation of the current policy}. 
The value estimate $Q(s,a)$ is affected by extrapolation error during a value update where the target policy selects an unfamiliar action $a'$ at the next state $s'$ in the backed-up value estimate, such that $(s',a')$ is unlikely, or not contained, in the dataset. 
The cause of extrapolation error can be attributed to several related factors:

\textbf{Absent Data.} If any state-action pair $(s,a)$ is unavailable, then error is introduced as some function of the amount of similar data and approximation error. This means that the estimate of $Q_\ta(s',\pi(s'))$ may be arbitrarily bad without sufficient data near $(s',\pi(s'))$.


\textbf{Model Bias.} When performing off-policy Q-learning with a batch $\B$, the Bellman operator $\mathcal{T}^\pi$ is approximated by sampling transitions tuples $(s,a,r,s')$ from $\B$ to estimate the expectation over $s'$. However, for a stochastic MDP, without infinite state-action visitation, this produces a biased estimate of the transition dynamics: 
\begin{equation}
\mathcal{T}^\pi Q(s,a) \approx \E_{s' \sim \B} [r + \y Q(s',\pi(s'))],
\end{equation}
where the expectation is with respect to transitions in the batch $\B$, rather than the true MDP. 

\textbf{Training Mismatch.} Even with sufficient data, in deep Q-learning systems, transitions are sampled uniformly from the dataset, giving a loss weighted with respect to the likelihood of data in the batch: 
\begin{equation} \label{eqn:DQN_loss}
\approx \frac{1}{|\B|} \sum_{(s,a,r,s')\in\B} ||r + \y Q_{\ta'}(s',\pi(s')) - Q_\ta(s,a) ||^2. 
\end{equation}
If the distribution of data in the batch does not correspond with the distribution under the current policy, the value function may be a poor estimate of actions selected by the current policy, due to the mismatch in training. 

We remark that re-weighting the loss in Equation~(\ref{eqn:DQN_loss}) with respect to the likelihood under the current policy can still result in poor estimates if state-action pairs with high likelihood under the current policy are not found in the batch. This means only a subset of possible policies can be evaluated accurately. As a result, learning a value estimate with off-policy data can result in large amounts of extrapolation error if the policy selects actions which are not similar to the data found in the batch. In the following section, we discuss how state of the art off-policy deep reinforcement learning algorithms fail to address the concern of extrapolation error, and demonstrate the implications in practical examples.

\subsection{Extrapolation Error in Deep Reinforcement Learning} \label{sec:drl_extrapolation}

Deep Q-learning algorithms \cite{DQN} have been labeled as off-policy due to their connection to off-policy Q-learning \cite{watkins1989qlearning}. However, these algorithms tend to use near-on-policy exploratory policies, such as $\e$-greedy, in conjunction with a replay buffer \cite{expreplay1992}. As a result, the generated dataset tends to be heavily correlated to the current policy. In this section, we examine how these off-policy algorithms perform when learning with uncorrelated datasets. Our results demonstrate that the performance of a state of the art deep actor-critic algorithm, DDPG \cite{DDPG}, deteriorates rapidly when the data is uncorrelated and the value estimate produced by the deep Q-network diverges. These results suggest that off-policy deep reinforcement learning algorithms are ineffective when learning \textit{truly off-policy}. 


Our practical experiments examine three different batch settings in OpenAI gym's Hopper-v1 environment 
\citep{mujoco,OpenAIGym}, which we use to train an off-policy DDPG agent with no interaction with the environment. Experiments with additional environments and specific details can be found in the Supplementary Material. 

\begin{figure}[t]
\centering
\captionsetup[subfloat]{captionskip=-8pt, justification=centering}
\includegraphics[width=\linewidth]{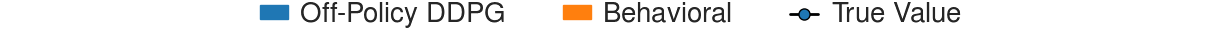}
\vspace{-4mm}

\includegraphics[trim={0 0 0 6mm}, clip, width=\linewidth]{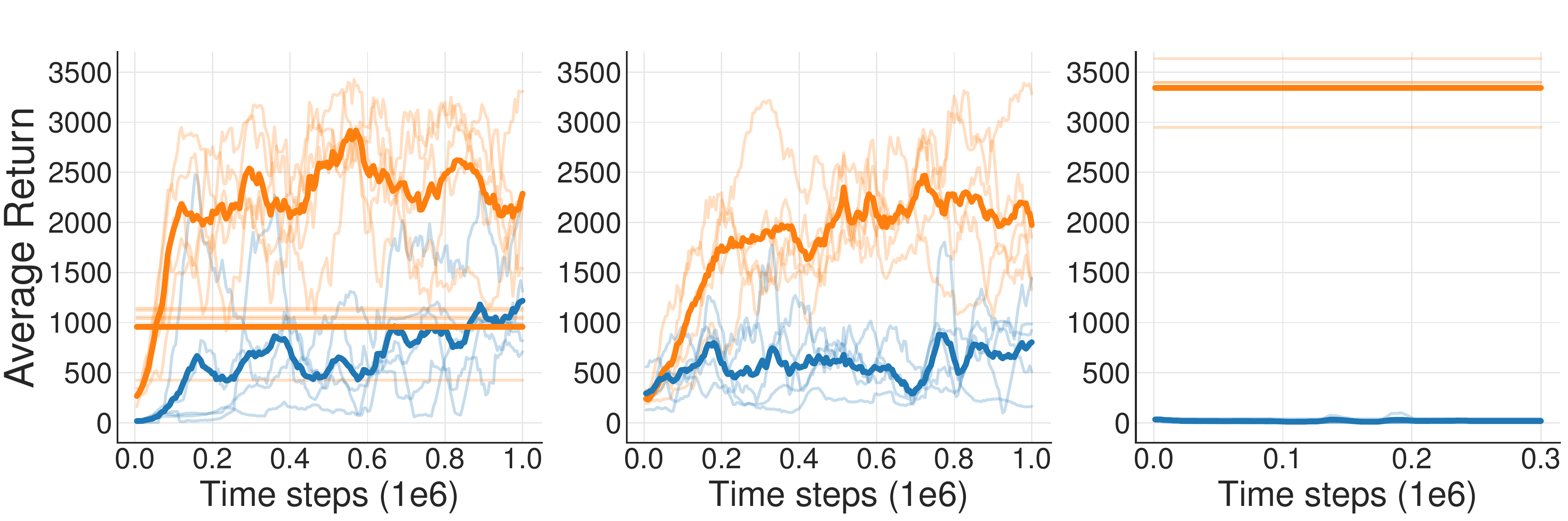}

\subfloat[Final buffer \newline performance]{\hspace{0.38\linewidth}}
\subfloat[Concurrent \newline performance]{\hspace{0.31\linewidth}}
\subfloat[Imitation \newline performance]{\hspace{0.31\linewidth}}

\includegraphics[trim={0 0 0 5mm}, clip, width=\linewidth]{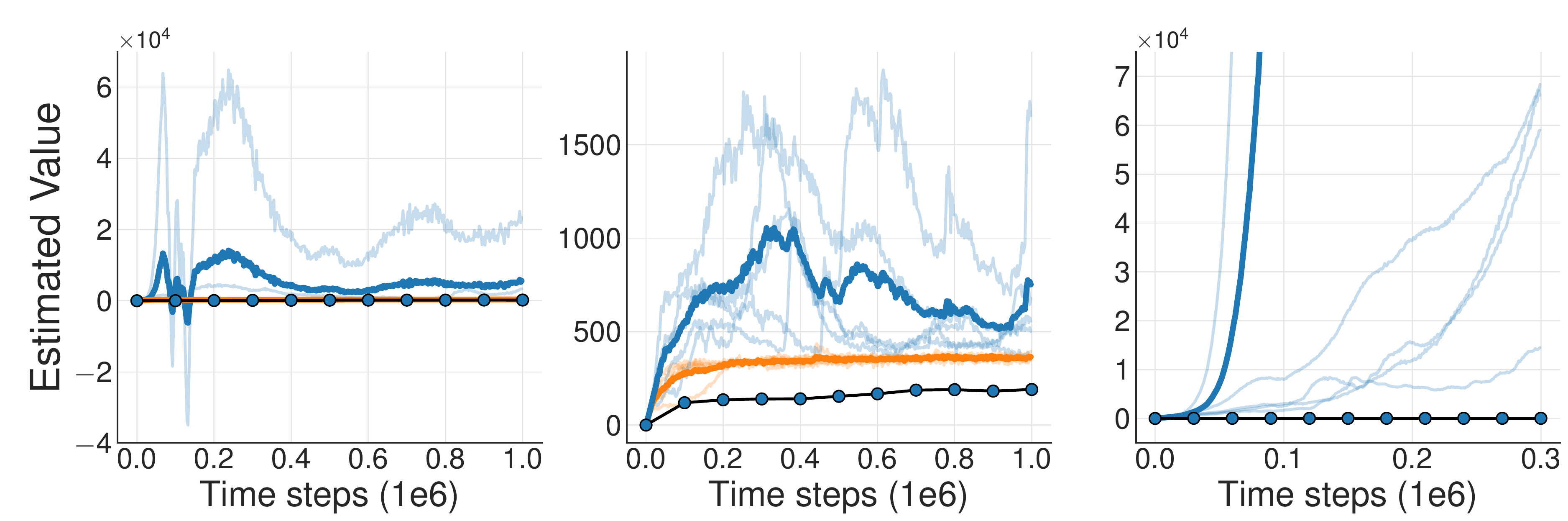}

\subfloat[Final buffer \newline value estimate]{\hspace{0.38\linewidth}}
\subfloat[Concurrent \newline value estimate]{\hspace{0.31\linewidth}}
\subfloat[Imitation \newline value estimate]{\hspace{0.31\linewidth}}

\caption{We examine the performance (top row) and corresponding value estimates (bottom row) of DDPG in three batch tasks on Hopper-v1. Each individual trial is plotted with a thin line, with the mean in bold (evaluated without exploration noise). Straight lines represent the average return of episodes contained in the batch (with exploration noise). An estimate of the true value of the off-policy agent, evaluated by Monte Carlo returns, is marked by a dotted line. In all three experiments, we observe a large gap in the performance between the behavioral and off-policy agent, even when learning from the same dataset (\textit{concurrent}). Furthermore, the value estimates are unstable or divergent across all tasks.}
\label{fig:extrapolation}
\vspace{-3mm}
\end{figure}

\textbf{Batch 1 (Final buffer).} We train a DDPG agent for 1 million time steps, adding $\N(0,0.5)$ Gaussian noise to actions for high exploration, and store all experienced transitions. This collection procedure creates a dataset with a diverse set of states and actions, with the aim of sufficient coverage. 

\textbf{Batch 2 (Concurrent).} We concurrently train the off-policy and behavioral DDPG agents, for 1 million time steps. To ensure sufficient exploration, a standard $\N(0,0.1)$ Gaussian noise is added to actions taken by the behavioral policy. 
Each transition experienced by the behavioral policy 
is stored in a buffer replay, which both agents learn from. As a result, both agents are trained with the identical dataset. 

\textbf{Batch 3 (Imitation).} A trained DDPG agent acts as an expert, and is used to collect a dataset of 1 million transitions.

In Figure~\ref{fig:extrapolation}, we graph the performance of the agents as they train with each batch, as well as their value estimates. Straight lines represent the average return of episodes contained in the batch. Additionally, we graph the learning performance of the behavioral agent for the relevant tasks. 

Our experiments demonstrate several surprising facts about off-policy deep reinforcement learning agents. In each task, the off-policy agent performances significantly worse than the behavioral agent. Even in the \textit{concurrent} experiment, where both agents are trained with the same dataset, there is a large gap in performance in every single trial. This result suggests that differences in the state distribution under the initial policies is enough for extrapolation error to drastically offset the performance of the off-policy agent. Additionally, the corresponding value estimate exhibits divergent behavior, while the value estimate of the behavioral agent is highly stable. In the \textit{final buffer} experiment, the off-policy agent is provided with a large and diverse dataset, with the aim of providing sufficient coverage of the initial policy. Even in this instance, the value estimate is highly unstable, and the performance suffers. In the \textit{imitation} setting, the agent is provided with expert data. However, the agent quickly learns to take non-expert actions, under the guise of optimistic extrapolation. As a result, the value estimates rapidly diverge and the agent fails to learn. 

Although extrapolation error is not necessarily positively biased, when combined with maximization in reinforcement learning algorithms, extrapolation error provides a source of noise that can induce a persistent overestimation bias~\cite{thrun1993bias,DoubleDQN,fujimoto2018addressing}. In an on-policy setting, extrapolation error may be a source of beneficial exploration through an implicit ``optimism in the face of uncertainty'' strategy \cite{lai1985asymptotically, jaksch2010near}. In this case, if the value function overestimates an unknown state-action pair, the policy will collect data in the region of uncertainty, and the value estimate will be corrected. However, when learning off-policy, or in a batch setting, extrapolation error will never be corrected due to the inability to collect new data. 


These experiments show extrapolation error can be highly detrimental to learning off-policy in a batch reinforcement learning setting. While the continuous state space and multi-dimensional action space in MuJoCo environments are contributing factors to extrapolation error, the scale of these tasks is small compared to real world settings. As a result, even with a sufficient amount of data collection, extrapolation error may still occur due to the concern of catastrophic forgetting \cite{mccloskey1989catastrophic, goodfellow2013empirical}. Consequently, off-policy reinforcement learning algorithms used in the real-world will require practical guarantees without exhaustive amounts of data. 


\section{Batch-Constrained Reinforcement Learning} \label{sec:DRL}

Current off-policy deep reinforcement learning algorithms fail to address extrapolation error by selecting actions with respect to a learned value estimate, without consideration of the accuracy of the estimate. As a result, certain out-of-distribution actions can be erroneously extrapolated to higher values. However, the value of an off-policy agent \textit{can} be accurately evaluated in regions where data is available. We propose a conceptually simple idea: to avoid extrapolation error \textit{a policy should induce a similar state-action visitation to the batch}. We denote policies which satisfy this notion as \textit{batch-constrained}. To optimize off-policy learning for a given batch, batch-constrained policies are trained to select actions with respect to three objectives:
\begin{enumerate}[topsep=0pt, nosep, label=(\arabic*)]
\item Minimize the distance of selected actions to the data in the batch.
\item Lead to states where familiar data can be observed.
\item Maximize the value function.
\end{enumerate}
We note the importance of objective (1) above the others, as the value function and estimates of future states may be arbitrarily poor without access to the corresponding transitions. That is, we cannot correctly estimate (2) and (3) unless (1) is sufficiently satisfied. 
As a result, we propose optimizing the value function, along with some measure of future certainty, with a constraint limiting the distance of selected actions to the batch. This is achieved in our deep reinforcement learning algorithm through a state-conditioned generative model, to produce likely actions under the batch. This generative model is combined with a network which aims to optimally perturb the generated actions in a small range, along with a Q-network, used to select the highest valued action. Finally, we train a pair of Q-networks, and take the minimum of their estimates during the value update. This update penalizes states which are unfamiliar, and pushes the policy to select actions which lead to certain data. 


We begin by analyzing the theoretical properties of batch-constrained policies in a finite MDP setting, where we are able to quantify extrapolation error precisely. We then introduce our deep reinforcement learning algorithm in detail, Batch-Constrained deep Q-learning (BCQ) by drawing inspiration from the tabular analogue.   



\subsection{Addressing Extrapolation Error in Finite MDPs} \label{sec:MDP}

In the finite MDP setting, extrapolation error can be described by the bias from the mismatch between the transitions contained in the buffer and the true MDP. 
We find that by inducing a data distribution that is contained entirely within the batch, batch-constrained policies can eliminate extrapolation error entirely for deterministic MDPs. 
In addition, we show that the batch-constrained variant of Q-learning converges to the optimal policy under the same conditions as the standard form of Q-learning. Moreover, we prove that for a deterministic MDP, batch-constrained Q-learning is guaranteed to match, or outperform, the behavioral policy when starting from any state contained in the batch. All of the proofs for this section can be found in the Supplementary Material. 


A value estimate $Q$ can be learned using an experience replay buffer $\B$. This involves sampling transition tuples $(s,a,r,s')$ with uniform probability, and applying the temporal difference update \cite{sutton1988tdlearning, watkins1989qlearning}:
\begin{equation}
Q(s,a) \leftarrow (1 - \al) Q(s,a) + \al \lp r + \y Q(s',\pi(s')) \rp.
\end{equation}
If $\pi(s') = \argmax_{a'} Q(s',a')$, this is known as Q-learning. 
Assuming a non-zero probability of sampling any possible transition tuple from the buffer and infinite updates, Q-learning converges to the optimal value function. 

We begin by showing that the value function $Q$ learned with the batch $\B$ corresponds to the value function for an alternate MDP $M_\B$. From the true MDP $M$ and initial values $Q(s,a)$, we define the new MDP $M_\B$ with the same action and state space as $M$, along with an additional terminal state $s_\text{init}$. $M_\B$ has transition probabilities $p_\B(s'|s,a) = \frac{N(s,a,s')}{\sum_{\tilde s} N(s,a,\tilde s)}$, where $N(s,a,s')$ is the number of times the tuple $(s,a,s')$ is observed in $\B$. If $\sum_{\tilde s} N(s,a,\tilde s) = 0$, then $p_\B(s_\text{init}|s,a) = 1$, where $r(s,a,s_\text{init})$ is set to the initialized value of $Q(s,a)$. 

\textbf{Theorem 1.} \textit{Performing Q-learning by sampling from a batch $\B$ converges to the optimal value function under the MDP $M_\B$.} 


We define $\e_\text{MDP}$ as the tabular extrapolation error, which accounts for the discrepancy between the value function $Q^\pi_\B$ computed with the batch $\B$ and the value function $Q^\pi$ computed with the true MDP $M$:
\begin{equation} \label{MDP}
    \e_{\text{MDP}}(s,a) = Q^\pi(s,a) - Q^\pi_\B(s,a).
\end{equation}
For any policy $\pi$, the exact form of $\e_{\text{MDP}}(s,a)$ can be computed through a Bellman-like equation:  
\begin{equation}
\begin{split}
\e_{\text{MDP}}(s,a)~&= \sum_{s'} \lp p_M(s'|s,a) - p_\B(s'|s,a) \rp \\
&\lp r(s,a,s') + \y \sum_{a'} \pi(a'|s') Q^\pi_\B(s',a') \rp \\
&+ p_M(s'|s,a) \y \sum_{a'} \pi(a'|s') \e_{\text{MDP}}(s',a').
\end{split}
\end{equation}
This means extrapolation error is a function of divergence in the transition distributions, weighted by value, along with the error at succeeding states. If the policy is chosen carefully, the error between value functions can be minimized by visiting regions where the transition distributions are similar. For simplicity, we denote 
\begin{equation}
    \e^\pi_{\text{MDP}} = \sum_s \mu_\pi(s) \sum_a \pi(a|s) |\e_{\text{MDP}}(s,a)|.
\end{equation}
To evaluate a policy $\pi$ exactly at relevant state-action pairs, only $\e^\pi_{\text{MDP}} = 0$ is required. We can then determine the condition required to evaluate the exact expected return of a policy without extrapolation error.

\textbf{Lemma 1.} \textit{For all reward functions, $\e^\pi_{\text{MDP}} = 0$ if and only if $p_\B(s'|s,a) = p_M(s'|s,a)$ for all $s' \in \mathcal{S}$ and $(s,a)$ such that $\mu_\pi(s) > 0$ and $\pi(a|s) > 0$.}

Lemma 1 states that if $M_\B$ and $M$ exhibit the same transition probabilities in regions of relevance, the policy can be accurately evaluated. For a stochastic MDP this may require an infinite number of samples to converge to the true distribution, however, for a deterministic MDP this requires only a single transition. 
This means a policy which only traverses transitions contained in the batch, can be evaluated without error. 
More formally, we denote a policy $\pi \in \Pi_\B$ as \textit{batch-constrained} if for all $(s,a)$ where $\mu_\pi(s) >0$ and $\pi(a|s) > 0$ then $(s,a) \in \B$. 
Additionally, we define a batch $\B$ as \textit{coherent} if for all $(s,a,s') \in \B$ then $s' \in \B$ unless $s'$ is a terminal state. This condition is trivially satisfied if the data is collected in trajectories, or if all possible states are contained in the batch. With a coherent batch, we can guarantee the existence of a batch-constrained policy. 

\textbf{Theorem 2.} \textit{For a deterministic MDP and all reward functions, $\e^\pi_{\text{MDP}} = 0$ if and only if the policy $\pi$ is batch-constrained. Furthermore, if $\B$ is coherent, then such a policy must exist if the start state $s_0 \in \B$.} 

Batch-constrained policies can be used in conjunction with Q-learning to form batch-constrained Q-learning (BCQL), which follows the standard tabular Q-learning update while constraining the possible actions with respect to the batch: 
\begin{equation}
Q(s,a) \leftarrow (1 - \al) Q(s,a) + \al (r + \y \max_{a' \text{s.t.} (s',a') \in \B} Q(s',a')).
\end{equation}
BCQL converges under the same conditions as the standard form of Q-learning, noting the batch-constraint is nonrestrictive given infinite state-action visitation. 

\textbf{Theorem 3.} \textit{Given the Robbins-Monro stochastic convergence conditions on the learning rate $\al$, and standard sampling requirements from the environment, BCQL converges to  the optimal value function $Q^*$.}

The more interesting property of BCQL is that for a deterministic MDP and any coherent batch $\B$, BCQL converges to the optimal batch-constrained policy $\pi^* \in \Pi_\B$ such that $Q^{\pi^*}(s,a) \geq Q^\pi(s,a)$ for all $\pi \in \Pi_\B$ and $(s,a) \in \B$. 

\textbf{Theorem 4.} \textit{Given a deterministic MDP and coherent batch $\B$, along with the Robbins-Monro stochastic convergence conditions on the learning rate $\al$ and standard sampling requirements on the batch $\B$, BCQL converges to $Q^\pi_\B(s,a)$ where $\pi^*(s) = \argmax_{a \text{ s.t.} (s,a) \in \B} Q^\pi_\B(s,a)$ is the optimal batch-constrained policy.}

This means that BCQL is guaranteed to outperform any behavioral policy when starting from any state contained in the batch, effectively outperforming imitation learning. Unlike standard Q-learning, there is no condition on state-action visitation, other than coherency in the batch. 

\subsection{Batch-Constrained Deep Reinforcement Learning} \label{sec:BCQ}

We introduce our approach to off-policy batch reinforcement learning, Batch-Constrained deep Q-learning (BCQ). BCQ approaches the notion of batch-constrained through a generative model. For a given state, BCQ generates plausible candidate actions with high similarity to the batch, and then selects the highest valued action through a learned Q-network. 
Furthermore, we bias this value estimate to penalize rare, or unseen, states through a modification to Clipped Double Q-learning \cite{fujimoto2018addressing}. As a result, BCQ learns a policy with a similar state-action visitation to the data in the batch, as inspired by the theoretical benefits of its tabular counterpart.

To maintain the notion of batch-constraint, we define a similarity metric by making the assumption that for a given state $s$, the similarity between $(s,a)$ and the state-action pairs in the batch $\B$ can be modelled using a learned state-conditioned marginal likelihood $P_\B^G(a|s)$. In this case, it follows that the policy maximizing $P_\B^G(a|s)$ would minimize the error induced by extrapolation from distant, or unseen, state-action pairs, by only selecting the most likely actions in the batch with respect to a given state. 
Given the difficulty of estimating $P_\B^G(a|s)$ in high-dimensional continuous spaces, 
we instead train a parametric generative model of the batch $G_\w(s)$, which we can sample actions from, as a reasonable approximation to $\argmax_a P_\B^G(a|s)$. 


For our generative model we use a conditional variational auto-encoder (VAE) \citep{kingma2013auto, sohn2015learning}, which models the distribution by transforming an underlying latent space\footnote{See the Supplementary Material for an introduction to VAEs.}. 
The generative model $G_\w$, alongside the value function $Q_\ta$, can be used as a policy by sampling $n$ actions from $G_\w$ and selecting the highest valued action according to the value estimate $Q_\ta$. 
To increase the diversity of seen actions, we introduce a perturbation model $\xi_\phi(s,a, \Phi)$, which outputs an adjustment to an action $a$ in the range $[-\Phi, \Phi]$. This enables access to actions in a constrained region, without having to sample from the generative model a prohibitive number of times. This results in the policy $\pi$: 
\begin{equation}
\begin{split}
\pi(s) = &\argmax_{a_i + \xi_\phi(s, a_i, \Phi)} Q_\ta(s,a_i + \xi_\phi(s, a_i, \Phi)),\\ 
&\{a_i \sim G_\w(s)\}_{i=1}^n.    
\end{split}
\end{equation}
The choice of $n$ and $\Phi$ creates a trade-off between an imitation learning and reinforcement learning algorithm. If $\Phi = 0$, and the number of sampled actions $n=1$, then the policy resembles behavioral cloning and as $\Phi \rightarrow a_{\text{max}} - a_{\text{min}}$ and $n \rightarrow \infty$, then the algorithm approaches Q-learning, as the policy begins to greedily maximize the value function over the entire action space. 

\begin{algorithm}[t]
  \caption{BCQ}
  \label{algorithm:BCQ}
\begin{algorithmic}
	\STATE \textbf{Input:} Batch $\B$, horizon $T$, target network update rate $\tau$, mini-batch size $N$, max perturbation $\Phi$, number of sampled actions $n$, minimum weighting $\lambda$. 
	\STATE Initialize Q-networks $Q_{\ta_1}, Q_{\ta_2}$, perturbation network $\xi_\phi$, and VAE $G_\w = \{E_{\w_1}, D_{\w_2}\}$, with random parameters $\ta_1$, $\ta_2$, $\phi$, $\w$, and target networks $Q_{\ta'_1}, Q_{\ta'_2}$, $\xi_{\phi'}$ with $\ta'_1 \leftarrow \ta_1, \ta'_2 \leftarrow \ta_2$, $\phi' \leftarrow \phi$.
	\FOR{$t=1$ {\bfseries to} $T$}
	\STATE Sample mini-batch of $N$ transitions $(s, a, r, s')$ from $\B$
	\STATE $\mu, \s = E_{\w_1}(s,a), \quad \tilde a = D_{\w_2}(s,z), \quad z \sim \N(\mu, \s)$ 
	\STATE $\w \leftarrow \argmin_\w \sum (a - \tilde a)^2 + D_{\text{KL}}(\N(\mu, \s)||\N(0,1))$ 
	\STATE Sample $n$ actions: $\{a_i \sim G_\w(s')\}_{i=1}^n$ 
	\STATE Perturb each action: $\{a_i = a_i + \xi_\phi(s',a_i, \Phi)\}_{i=1}^n$
	\STATE Set value target $y$ (Eqn. \ref{eqn:soft_min})
	\STATE $\ta \leftarrow \argmin_\ta \sum (y - Q_\ta(s,a))^2$ 
	\STATE $\phi \leftarrow \argmax_\phi \sum Q_{\ta_1}(s,a + \xi_\phi(s, a, \Phi)), a \sim G_\w(s)$ 
	\STATE Update target networks: $\ta'_i \leftarrow \tau \ta + (1 - \tau) \ta'_i$
	\STATE $\phi' \leftarrow \tau \phi + (1 - \tau) \phi'$
	\ENDFOR
\end{algorithmic}
\end{algorithm}

The perturbation model $\xi_\phi$ can be trained to maximize $Q_\ta(s,a)$ through the deterministic policy gradient algorithm \citep{DPG} by sampling $a \sim G_\w(s)$: 
\begin{equation}
\phi \leftarrow \argmax_{\phi} \sum_{(s,a) \in \B} Q_\ta(s,a + \xi_{\phi}(s, a, \Phi)). 
\end{equation}


To penalize uncertainty over future states, we modify Clipped Double Q-learning \cite{fujimoto2018addressing}, which estimates the value by taking the minimum between two Q-networks $\{ Q_{\ta_1}, Q_{\ta_2} \}$. Although originally used as a countermeasure to overestimation bias \cite{thrun1993bias, hasselt2010double}, the minimum operator also penalizes high variance estimates in regions of uncertainty, and pushes the policy to favor actions which lead to states contained in the batch. In particular, we take a convex combination of the two values, with a higher weight on the minimum, to form a learning target which is used by both Q-networks:
\vspace{-3mm}
\begin{equation} \label{eqn:soft_min}
r + \y \max_{a_i} \lb \lambda \min_{j=1,2} Q_{\ta'_j}(s',a_i) + (1 - \lambda) \max_{j=1,2} Q_{\ta'_j}(s',a_i) \rb
\end{equation}
where $a_i$ corresponds to the perturbed actions, sampled from the generative model. If we set $\lambda = 1$, this update corresponds to Clipped Double Q-learning. We use this weighted minimum as the constrained updates produces less overestimation bias than a purely greedy policy update, and enables control over how heavily uncertainty at future time steps is penalized through the choice of $\lambda$.  


This forms Batch-Constrained deep Q-learning (BCQ), which maintains four parametrized networks: a generative model $G_\w(s)$, a perturbation model $\xi_\phi(s,a)$, and two Q-networks $Q_{\ta_1}(s,a), Q_{\ta_2}(s,a)$. We summarize BCQ in Algorithm~\ref{algorithm:BCQ}. 
In the following section, we demonstrate BCQ results in stable value learning and a strong performance in the batch setting. Furthermore, we find that only a single choice of hyper-parameters is necessary for a wide range of tasks and environments.

\section{Experiments}\label{sec:exp}

\begin{figure}[t]
\centering
\includegraphics[width=1\linewidth]{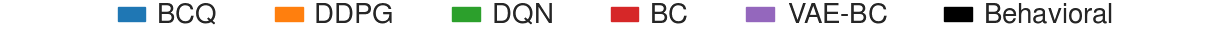} 
\vspace{-6mm}

\subfloat[Final buffer performance]{\includegraphics[width=1\linewidth]{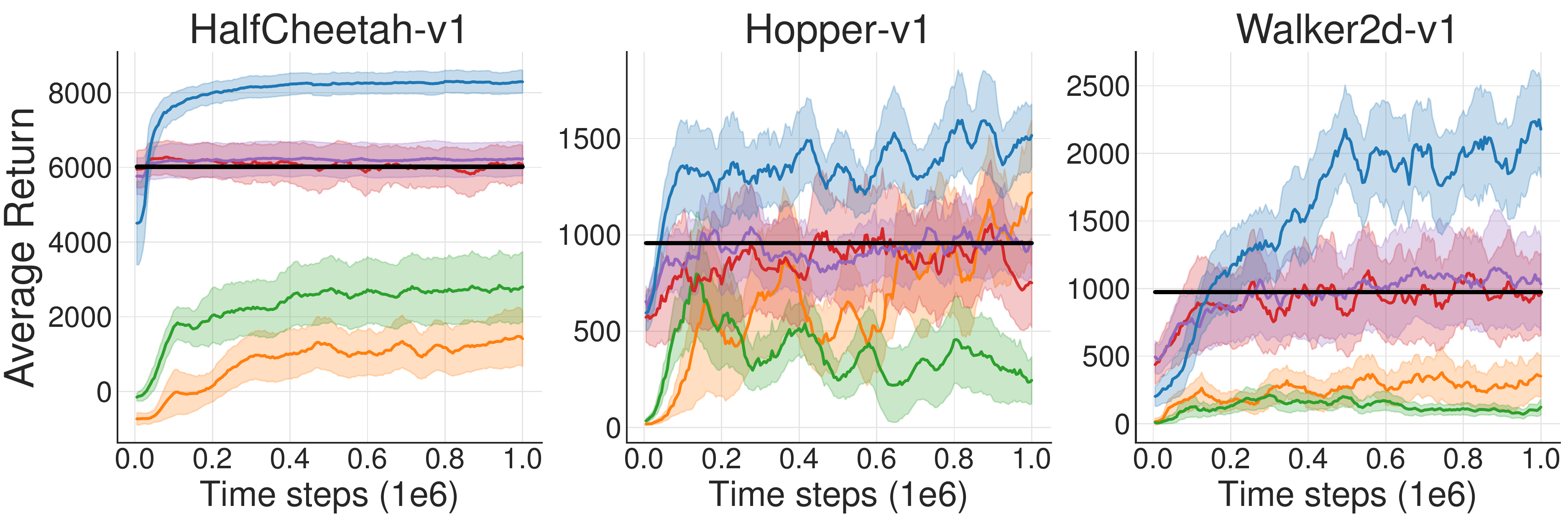}}
\vspace{-3mm}
\subfloat[Concurrent performance]{\includegraphics[width=1\linewidth]{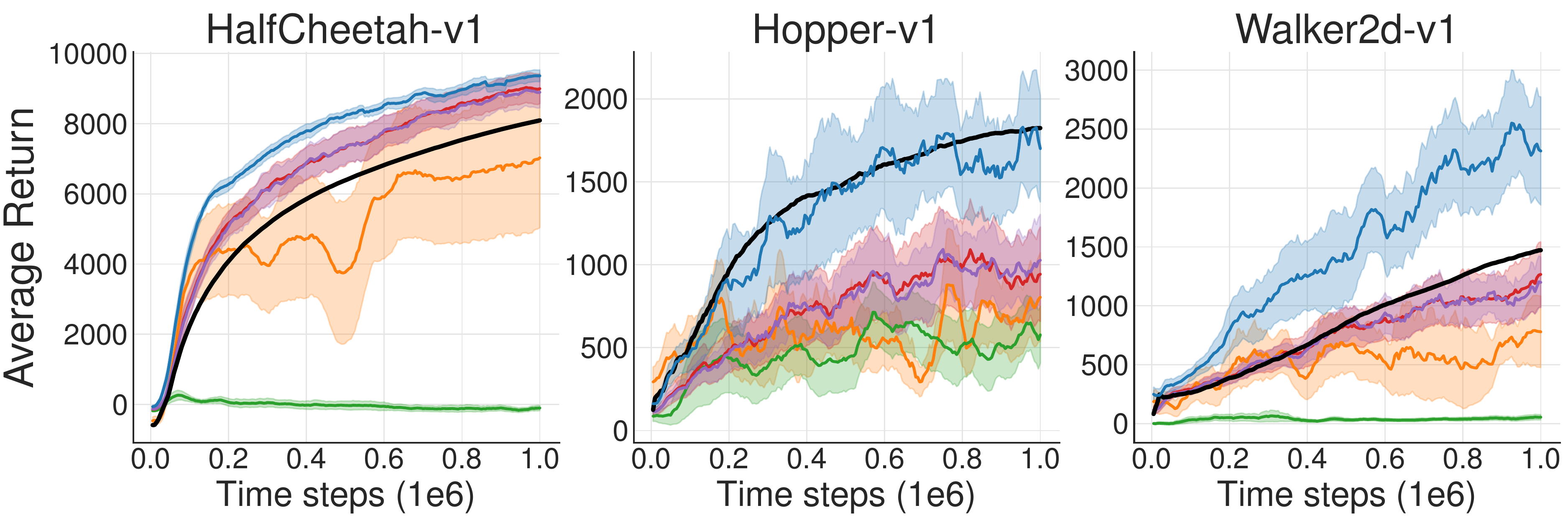}}
\vspace{-3mm}
\subfloat[Imitation performance]{\includegraphics[width=1\linewidth]{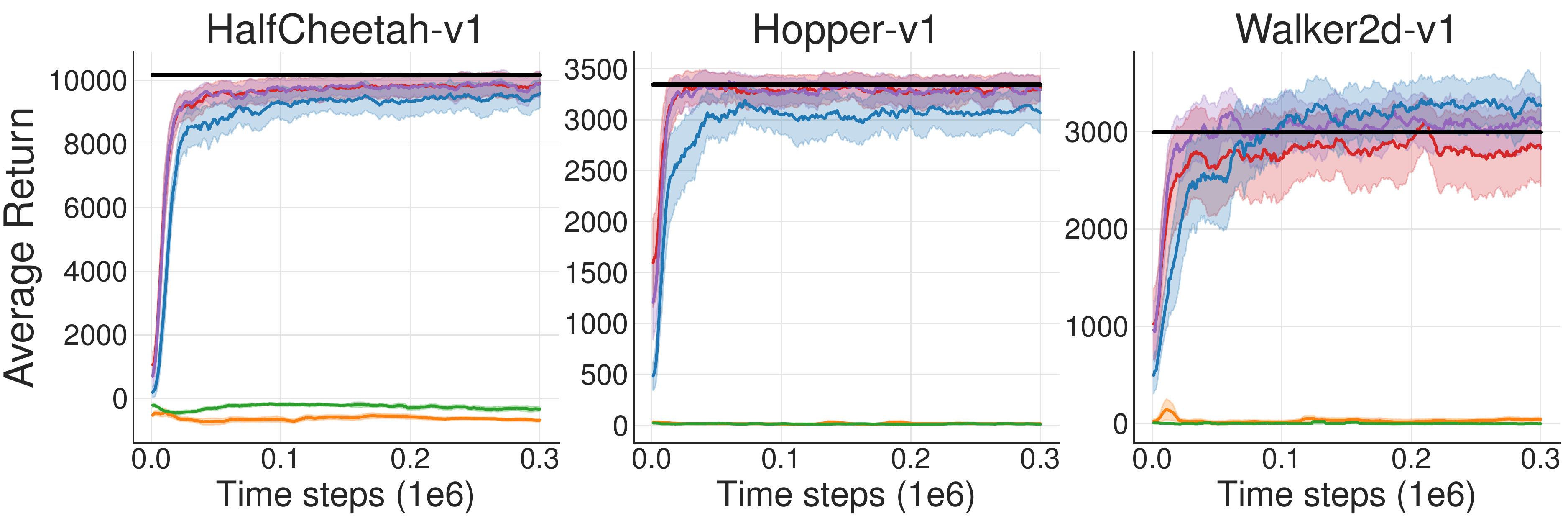}}
\vspace{-3mm}
\subfloat[Imperfect demonstrations performance]{\includegraphics[width=1\linewidth]{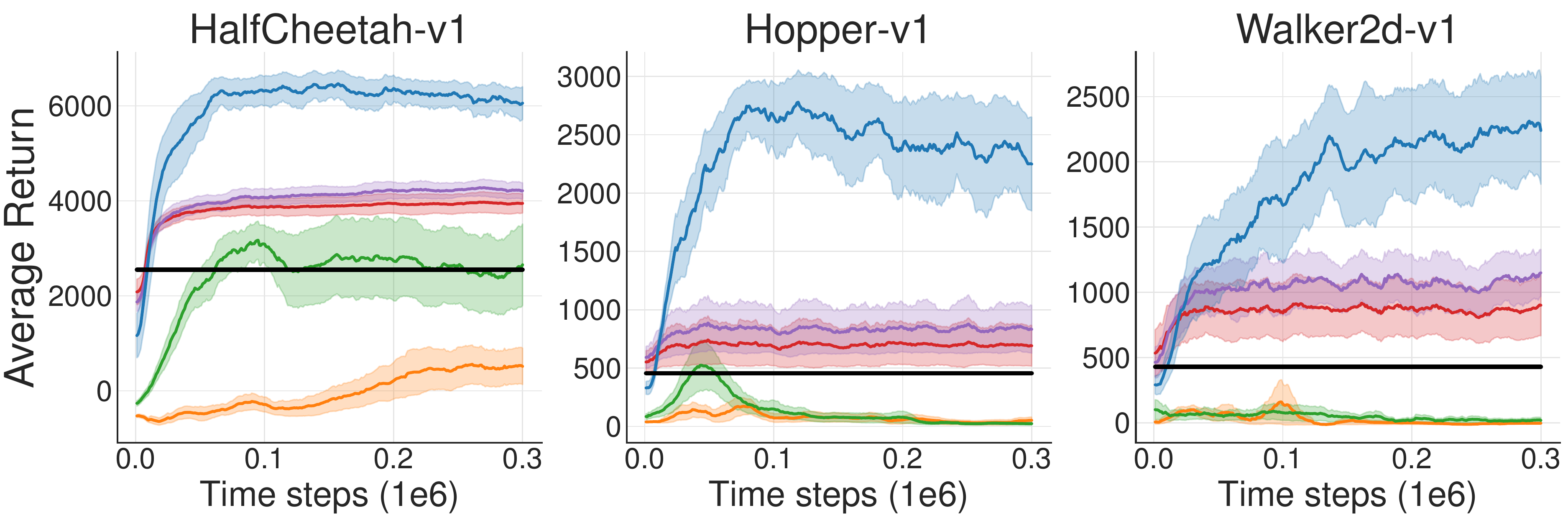}}
\vspace{-2mm}
\caption{We evaluate BCQ and several baselines on the experiments from Section \ref{sec:drl_extrapolation}, as well as the imperfect demonstrations task. The shaded area represents half a standard deviation. 
The bold black line measures the average return of episodes contained in the batch. 
Only BCQ matches or outperforms the performance of the behavioral policy in all  tasks.}
\label{results:extrapolation}
\vspace{-4mm}
\end{figure}


To evaluate the effectiveness of Batch-Constrained deep Q-learning (BCQ) in a high-dimensional setting, we focus on MuJoCo environments in OpenAI gym \citep{mujoco,OpenAIGym}. For reproducibility, we make no modifications to the original environments or reward functions. We compare our method with DDPG \citep{DDPG}, DQN \citep{DQN} using an independently discretized action space, a feed-forward behavioral cloning method (BC), and a variant with a VAE (VAE-BC), using $G_\w(s)$ from BCQ. Exact implementation and experimental details are provided in the Supplementary Material.

\begin{figure}[t]
\centering
\captionsetup[subfloat]{captionskip=-8pt}
\includegraphics[width=1\linewidth]{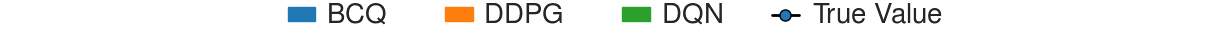} 

\includegraphics[trim={0 0 0 5mm}, clip, width=\linewidth]{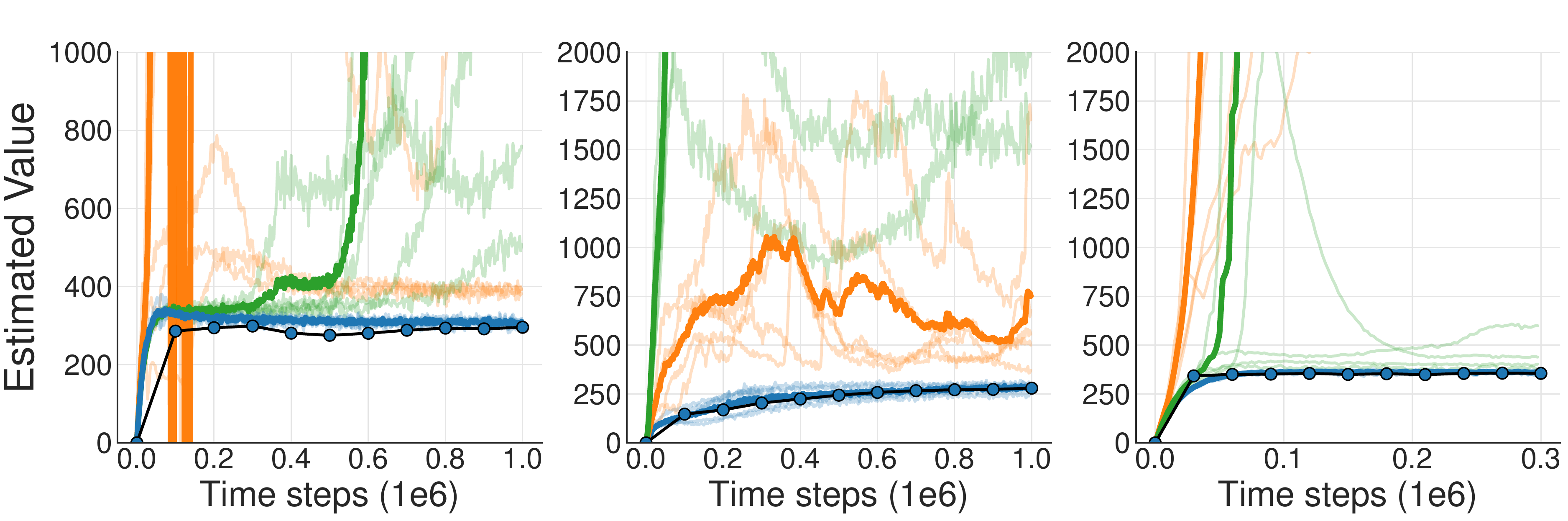}

\subfloat[Final Buffer]{\hspace{0.38\linewidth}}
\subfloat[Concurrent]{\hspace{0.31\linewidth}}
\subfloat[Imitation]{\hspace{0.31\linewidth}}
\vspace{-2mm}
\caption{We examine the value estimates of BCQ, along with DDPG and DQN on the experiments from Section \ref{sec:drl_extrapolation} in the Hopper-v1 environment. Each individual trial is plotted, with the mean in bold. An estimate of the true value of BCQ, evaluated by Monte Carlo returns, is marked by a dotted line. Unlike the state of the art baselines, BCQ exhibits a highly stable value function in each task. Graphs for the other environments and imperfect demonstrations task can be found in the Supplementary Material.} \label{results:values}
\vspace{-4mm}
\end{figure}

We evaluate each method following the three experiments defined in Section \ref{sec:drl_extrapolation}. In \textit{final buffer} the off-policy agents learn from the final replay buffer gathered by training a DDPG agent over a million time steps. In \textit{concurrent} the off-policy agents learn concurrently, with the same replay buffer, as the behavioral DDPG policy, and in \textit{imitation}, the agents learn from a dataset collected by an expert policy. Additionally, to study the robustness of BCQ to noisy and multi-modal data, we include an \textit{imperfect demonstrations} task, in which the agents are trained with a batch of 100k transitions collected by an expert policy, with two sources of noise. The behavioral policy selects actions randomly with probability $0.3$ and with high exploratory noise $\N(0,0.3)$ added to the remaining actions. The experimental results for these tasks are reported in Figure~\ref{results:extrapolation}. Furthermore, the estimated values of BCQ, DDPG and DQN, and the true value of BCQ are displayed in Figure~\ref{results:values}.

Our approach, BCQ, is the only algorithm which succeeds at all tasks, matching or outperforming the behavioral policy in each instance, and outperforming all other agents, besides in the imitation learning task where behavioral cloning unsurprisingly performs the best.
These results demonstrate that our algorithm can be used as a single approach for both imitation learning and off-policy reinforcement learning, with a single set of fixed hyper-parameters. 
Furthermore, unlike the deep reinforcement learning algorithms, DDPG and DQN, BCQ exhibits a highly stable value function in the presence of off-policy samples, suggesting extrapolation error has been successfully mitigated through the batch-constraint.  
In the imperfect demonstrations task, we find that both deep reinforcement learning and imitation learning algorithms perform poorly. BCQ, however, is able to strongly outperform the noisy demonstrator, disentangling poor and expert actions. Furthermore, compared to current deep reinforcement learning algorithms, which can require millions of time steps \cite{duan2016benchmark, hendersonRL2017}, BCQ attains a high performance in remarkably few iterations. This suggests our approach effectively leverages expert transitions, even in the presence of noise. 


\section{Related Work}

\textbf{Batch Reinforcement Learning.} While batch reinforcement learning algorithms have been shown to be convergent with non-parametric function approximators such as averagers \citep{gordon1995stable} and kernel methods \citep{ormoneit2002kernel}, they make no guarantees on the quality of the policy without infinite data. Other batch algorithms, such as fitted Q-iteration, have used other function approximators, including decision trees \citep{ernst2005tree} and neural networks \citep{riedmiller2005neural}, but come without convergence guarantees. 
Unlike many previous approaches to off-policy policy evaluation \cite{peshkin2002learning, thomas2015high, liu2018representation},
our work focuses on constraining the policy to a subset of policies which can be adequately evaluated, rather than the process of evaluation itself. Additionally, off-policy algorithms which rely on importance sampling \citep{precup2001off, jiang2016doubly, munos2016safe} may not be applicable in a batch setting, requiring access to the action probabilities under the behavioral policy, and scale poorly to multi-dimensional action spaces. 
Reinforcement learning with a replay buffer \citep{expreplay1992} can be considered a form of batch reinforcement learning, and is a standard tool for off-policy deep reinforcement learning algorithms \citep{DQN}. It has been observed that a large replay buffer can be detrimental to performance \citep{de2015expreplay,zhang2017expreplay} and the diversity of states in the buffer is an important factor for performance \citep{de2016improved}. \citet{isele2018selective} observed the performance of an agent was strongest when the distribution of data in the replay buffer matched the test distribution. These results defend the notion that extrapolation error is an important factor in the performance off-policy reinforcement learning. 

\textbf{Imitation Learning.} Imitation learning and its variants are well studied problems \citep{schaal1999imitation,argall2009survey, hussein2017imitation}. 
Imitation has been combined with reinforcement learning, via learning from demonstrations methods \cite{kim2013learning,piot2014boosted,chemali2015direct}, with deep reinforcement learning extensions \cite{hester2017deep,vevcerik2017leveraging}, and modified policy gradient approaches \cite{ho2016model, sun2017deeply, cheng2018fast, sun2018truncated}. While effective, these interactive methods are inadequate for batch reinforcement learning as they require either an explicit distinction between expert and non-expert data, further on-policy data collection or access to an oracle. 
Research in imitation, and inverse reinforcement learning, with robustness to noise is an emerging area \citep{evans2016learning, nair2018overcoming}, but relies on some form of expert data. \citet{gao2018imperfect} introduced an imitation learning algorithm which learned from imperfect demonstrations, by favoring seen actions, but is limited to discrete actions. Our work also connects to residual policy learning \cite{johannink2018residual, silver2018residual}, where the initial policy is the generative model, rather than an expert or feedback controller. 

\textbf{Uncertainty in Reinforcement Learning.} Uncertainty estimates in deep reinforcement learning have generally been used to encourage exploration \citep{dearden1998bayesian, strehl2008analysis, UncertaintyBellman, azizzadenesheli2018bayesian}. Other methods have examined approximating the Bayesian posterior of the value function \citep{osband2016deep, osband2018randomized, touati2018randomized}, again using the variance to encourage exploration to unseen regions of the state space. 
In model-based reinforcement learning, uncertainty has been used for exploration, but also for the opposite effect--to push the policy towards regions of certainty in the model. This is used to combat the well-known problems with compounding model errors, and is present in policy search methods \citep{deisenroth2011pilco, gal2016improving, higuera2018synthesizing, xu2018algorithmic}, or combined with trajectory optimization \citep{chua2018deep} or value-based methods \citep{buckman2018sample}. Our work connects to policy methods with conservative updates \citep{kakade2002approximately}, such as trust region \citep{trpo,achiam2017constrained,pham2018constrained} and information-theoretic methods \citep{peters2010relative,van2017non}, which aim to keep the updated policy similar to the previous policy. These methods avoid explicit uncertainty estimates, and rather force policy updates into a constrained range before collecting new data, limiting errors introduced by large changes in the policy. Similarly, our approach can be thought of as an off-policy variant, where the policy aims to be kept close, in output space, to any combination of the previous policies which performed data collection.

\section{Conclusion}

In this work, we demonstrate a critical problem in off-policy reinforcement learning with finite data, where the value target introduces error by including an estimate of unseen state-action pairs. This phenomenon, which we denote \textit{\mbox{extrapolation} \mbox{error}}, has important implications for off-policy and batch reinforcement learning, as it is generally implausible to have complete state-action coverage in any practical setting. We present batch-constrained reinforcement learning--acting close to on-policy with respect to the available data, as an answer to extrapolation error. When extended to a deep reinforcement learning setting, our algorithm, Batch-Constrained deep Q-learning (BCQ), is the first continuous control algorithm capable of learning from arbitrary batch data, without exploration. Due to the importance of batch reinforcement learning for practical applications, we believe BCQ will be a strong foothold for future algorithms to build on, while furthering our understanding of the systematic risks in Q-learning \cite{thrun1993bias, lu2018delusional}.


\bibliography{example_paper}
\bibliographystyle{icml2019}

\clearpage

\onecolumn
\icmltitle{Off-Policy Deep Reinforcement Learning without Exploration: \newline Supplementary Material}
\icmltitlerunning{Off-Policy Deep Reinforcement Learning without Exploration: Supplementary Material}
\appendix

\section{Missing Proofs} \label{appendix:proofs}

\subsection{Proofs and Details from Section 4.1}

\textbf{Definition 1.} We define a coherent batch $\B$ as a batch such that if $(s,a,s') \in \B$ then $s' \in \B$ unless $s'$ is a terminal state.

\textbf{Definition 2.} We define $\e_{\text{MDP}}(s,a) = Q^\pi(s,a) - Q^\pi_\B(s,a)$ as the error between the true value of a policy $\pi$ in the MDP $M$ and the value of $\pi$ when learned with a batch $\B$. 

\textbf{Definition 3.} For simplicity in notation, we denote
\begin{equation}
    \e^\pi_{\text{MDP}} = \sum_s \mu_\pi(s) \sum_a \pi(a|s) |\e_{\text{MDP}}(s,a)|.
\end{equation}
To evaluate a policy $\pi$ exactly at relevant state-action pairs, only $\e^\pi_{\text{MDP}} = 0$ is required. 

\textbf{Definition 4.} We define the optimal batch-constrained policy $\pi^* \in \Pi_\B$ such that $Q^{\pi^*}(s,a) \geq Q^\pi(s,a)$ for all $\pi \in \Pi_\B$ and $(s,a) \in \B$. 

\bigskip

\textbf{Algorithm 1.} Batch-Constrained Q-learning (BCQL) maintains a tabular value function $Q(s,a)$ for each possible state-action pair $(s,a)$. A transition tuple $(s,a,r,s')$ is sampled from the batch $\B$ with uniform probability and the following update rule is applied, with learning rate $\al$: 
\begin{equation}
Q(s,a) \leftarrow (1 - \al) Q(s,a) + \al (r + \y \max_{a' \text{s.t.} (s',a') \in \B} Q(s',a')).
\end{equation}


\textbf{Theorem 1.} \textit{Performing Q-learning by sampling from a batch $\B$ converges to the optimal value function under the MDP $M_\B$.} 

\textit{Proof.} Again, the MDP $M_\B$ is defined by the same action and state space as $M$, with an additional terminal state $s_\text{init}$. $M_\B$ has transition probabilities $p_\B(s'|s,a) = \frac{N(s,a,s')}{\sum_{\tilde s} N(s,a,\tilde s)}$, where $N(s,a,s')$ is the number of times the tuple $(s,a,s')$ is observed in $\B$. If $\sum_{\tilde s} N(s,a,\tilde s) = 0$, then $p_\B(s_\text{init}|s,a) = 1$, where $r(s_\text{init},s,a)$ is to the initialized value of $Q(s,a)$.

For any given MDP Q-learning converges to the optimal value function given infinite state-action visitation and some standard assumptions (see Section \ref{appendix:qlearning_proof}). Now note that sampling under a batch $\B$ with uniform probability satisfies the infinite state-action visitation assumptions of the MDP $M_\B$, where given $(s,a)$, the probability of sampling $(s,a,s')$ corresponds to $p(s'|s,a) = \frac{N(s,a,s')}{\sum_{\tilde s} N(s,a,\tilde s)}$ in the limit. We remark that for $(s,a) \notin \B$, $Q(s,a)$ will never be updated, and will return the initialized value, which corresponds to the terminal transition $s_\text{init}$. It follows that sampling from $\B$ is equivalent to sampling from the MDP $M_\B$, and Q-learning converges to the optimal value function under $M_\B$.




\bigskip

\textbf{Remark 1.} \textit{For any policy $\pi$ and state-action pair $(s,a)$, the error term $\e_{\text{MDP}}(s,a)$ satisfies the following Bellman-like equation:}
\begin{equation}
\begin{split}
\e_{\text{MDP}}(s,a) =~&\sum_{s'} \lp p_M(s'|s,a) - p_\B(s'|s,a) \rp \lp r(s,a,s') + \y \sum_{a'} \pi(a'|s') \lp Q^\pi_\B(s',a') \rp \rp \\
&+ p_M(s'|s,a) \y \sum_{a'} \pi(a'|s') \e_{\text{MDP}}(s',a').
\end{split}
\end{equation}

\textit{Proof.} Proof follows by expanding each $Q$, rearranging terms and then simplifying the expression. 
\begin{equation}
\begin{split}
\e_{\text{MDP}}(s,a) =&~Q^\pi(s,a) - Q^\pi_\B(s,a) \\
=& \sum_{s'} p_M(s'|s,a) \lp r(s,a,s') + \y \sum_{a'} \pi(a'|s') Q^\pi(s',a') \rp - Q^\pi_\B(s,a) \\
=& \sum_{s'} p_M(s'|s,a) \lp r(s,a,s') + \y \sum_{a'} \pi(a'|s') Q^\pi(s',a') \rp \\
&- \lp \sum_{s'} p_\B(s'|s,a) \lp r(s,a,s') +  \y \sum_{a'} \pi(a'|s') Q^\pi_\B(s',a') \rp \rp \\
=& \sum_{s'} \lp p_M(s'|s,a) - p_\B(s'|s,a) \rp r(s,a,s') + p_M(s'|s,a) \y \sum_{a'} \pi(a'|s') \lp Q^\pi_\B(s',a') + \e_{\text{MDP}}(s',a') \rp \\
&- p_\B(s'|s,a) \y \sum_{a'} \pi(a'|s') Q^\pi_\B(s',a') \\
%
=& \sum_{s'} \lp p_M(s'|s,a) - p_\B(s'|s,a) \rp r(s,a,s') + p_M(s'|s,a) \y \sum_{a'} \pi(a'|s') \lp Q^\pi_\B(s',a') + \e_{\text{MDP}}(s',a') \rp \\
&+ p_M(s'|s,a) \y \sum_{a'} \pi(a'|s') \lp \e_{\text{MDP}}(s',a') -\e_{\text{MDP}}(s',a') \rp  - p_\B(s'|s,a) \y \sum_{a'} \pi(a'|s') Q^\pi_\B(s',a') \\
=& \sum_{s'} \lp p_M(s'|s,a) - p_\B(s'|s,a) \rp \lp r(s,a,s') + \y \sum_{a'} \pi(a'|s') Q^\pi_\B(s',a') \rp \\
&+ p_M(s'|s,a) \y \sum_{a'} \pi(a'|s') \e_{\text{MDP}}(s',a')
\end{split}
\end{equation}

\bigskip

\textbf{Lemma 1.} \textit{For all reward functions, $\e^\pi_{\text{MDP}} = 0$ if and only if $p_\B(s'|s,a) = p_M(s'|s,a)$ for all $s' \in \mathcal{S}$ and $(s,a)$ such that $\mu_\pi(s) > 0$ and $\pi(a|s) > 0$.}

\textit{Proof.} From Remark 1, we note that the form of $\e_{\text{MDP}}(s,a)$, since no assumptions can be made on the reward function and therefore the expression $r(s,a,s') + \y \sum_{a'} \pi(a'|s') Q^\pi_\B(s',a')$, we have that $\e_{\text{MDP}}(s,a) = 0$ if and only if $p_\B(s'|s,a) = p_M(s'|s,a)$ for all $s' \in \mathcal{S}$ and $p_M(s'|s,a) \y \sum_{a'} \pi(a'|s') \e_{\text{MDP}}(s',a') = 0$. 

$(\Rightarrow)$ Now we note that if $\e_{\text{MDP}}(s,a) = 0$ then $p_M(s'|s,a) \y \sum_{a'} \pi(a'|s') \e_{\text{MDP}}(s',a') = 0$ by the relationship defined by Remark 1 and the condition on the reward function. It follows that we must have $p_\B(s'|s,a) = p_M(s'|s,a)$ for all $s' \in \mathcal{S}$.

$(\Leftarrow)$ If we have $\sum_{s'} |p_M(s'|s,a) - p_\B(s'|s,a)| = 0$ for all $(s,a)$ such that $\mu_\pi(s) > 0$ and $\pi(a|s) > 0$, then for any $(s,a)$ under the given conditions, we have $\e(s,a) = \sum_{s'} p_M(s'|s,a) \y \sum_{a'} \pi(a'|s')\e(s',a')$. Recursively expanding the $\e$ term, we arrive at $\e(s,a) = 0 + \y 0 + \y^2 0 + ... = 0$. 



\bigskip

\textbf{Theorem 2.} \textit{For a deterministic MDP and all reward functions, $\e^\pi_{\text{MDP}} = 0$ if and only if the policy $\pi$ is batch-constrained. Furthermore, if $\B$ is coherent, then such a policy must exist if the start state $s_0 \in \B$.}

\textit{Proof.} The first part of the Theorem follows from Lemma 1, noting that for a deterministic policy $\pi$, if $(s,a) \in \B$ then we must have $p_\B(s'|s,a) = p_M(s'|s,a)$ for all $s' \in \mathcal{S}$. 

We can construct the batch-constrained policy by selecting $a$ in the state $s \in \B$, such that $(s,a) \in \B$. Since the MDP is deterministic and the batch is coherent, when starting from $s_0$, we must be able to follow at least one trajectory until termination. 

\bigskip

\textbf{Theorem 3.} \textit{Given the Robbins-Monro stochastic convergence conditions on the learning rate $\al$, and standard sampling requirements from the environment, BCQL converges to  the optimal value function $Q^*$.}

\textit{Proof.} Follows from proof of convergence of Q-learning (see Section \ref{appendix:qlearning_proof}), noting the batch-constraint is non-restrictive with a batch which contains all possible transitions.

\bigskip

\textbf{Theorem 4.} \textit{Given a deterministic MDP and coherent batch $\B$, along with the Robbins-Monro stochastic convergence conditions on the learning rate $\al$ and standard sampling requirements on the batch $\B$, BCQL converges to $Q^\pi_\B(s,a)$ where $\pi^*(s) = \argmax_{a \text{ s.t.} (s,a) \in \B} Q^\pi_\B(s,a)$ is the optimal batch-constrained policy.}

\textit{Proof.} Results follows from Theorem 1, which states Q-learning learns the optimal value for the MDP $M_\B$ for state-action pairs in $(s,a)$. However, for a deterministic MDP $M_\B$ corresponds to the true MDP in all seen state-action pairs. Noting that batch-constrained policies operate only on state-action pairs where $M_\B$ corresponds to the true MDP, it follows that $\pi^*$ will be the optimal batch-constrained policy from the optimality of Q-learning. 


\bigskip 

\subsection{Sketch of the Proof of Convergence of Q-Learning} \label{appendix:qlearning_proof}

The proof of convergence of Q-learning relies large on the following lemma \cite{singh2000convergence}: 

\textbf{Lemma 2.} \textit{Consider a stochastic process $(\zeta_t, \Delta_t, F_t), t \geq 0$ where $\zeta_t, \Delta_t, F_t: X \rightarrow \mathbb{R}$ satisfy the equation:
\begin{equation}
\Delta_{t+1}(x_t) = (1 - \zeta_t(x_t))\Delta_t(x_t) + \zeta_t(x_t)F_t(x_t),
\end{equation}
where $x_t \in X$ and $t=0,1,2,...$. Let $P_t$ be a sequence of increasing $\sigma$-fields such that $\zeta_0$ and $\Delta_0$ are $P_0$-measurable and $\zeta_t, \Delta_t$ and $F_{t-1}$ are $P_t$-measurable, $t=1,2,...$. Assume that the following hold:
\begin{enumerate}[topsep=0pt]
\item The set $X$ is finite.
\item $\zeta_t(x_t) \in [0,1], \sum_t \zeta_t(x_t) = \infty, \sum_t (\zeta_t(x_t))^2 < \infty$ with probability $1$ and $\forall x \neq x_t : \zeta(x) = 0$.
\item $|| \E \lb F_t | P_t \rb || \leq \kappa || \Delta_t || + c_t$ where $\kappa \in [0,1)$ and $c_t$ converges to $0$ with probability $1$.
\item Var$\lb F_t(x_t)|P_t \rb \leq K(1 + \kappa||\Delta_t||)^2$, where $K$ is some constant 
\end{enumerate}
Where $||\cdot||$ denotes the maximum norm. Then $\Delta_t$ converges to $0$ with probability $1$.}

\bigskip

\textbf{Sketch of Proof of Convergence of Q-Learning.} We set $\Delta_t = Q_t(s,a) - Q^*(s,a)$. Then convergence follows by satisfying the conditions of Lemma 2. Condition 1 is satisfied by the finite MDP, setting $X = \mathcal{S} \times \mathcal{A}$. Condition 2 is satisfied by the assumption of Robbins-Monro stochastic convergence conditions on the learning rate $\al_t$, setting $\zeta_t = \al_t$. Condition 4 is satisfied by the bounded reward function, where $F_t(s,a) = r(s,a,s') + \y \max_{a'} Q(s', a') - Q^*(s,a)$, and the sequence $P_t = \{Q_0, s_0, a_0, \al_0, r_1, s_1, ...s_t, a_t\}$. Finally, Condition 3 follows from the contraction of the Bellman Operator $\mathcal{T}$, requiring infinite state-action visitation, infinite updates and $\y < 1$. 

Additional and more complete details can be found in numerous resources \cite{dayan1992q, singh2000convergence, melo2001convergence}. 

\clearpage

\section{Missing Graphs}

\subsection{Extrapolation Error in Deep Reinforcement Learning}

\begin{figure}[ht]
\centering
\includegraphics[width=0.65\linewidth]{extrapolation_clean/ext_legend.png}

\subfloat[Final buffer performance]{\includegraphics[width=0.5\linewidth]{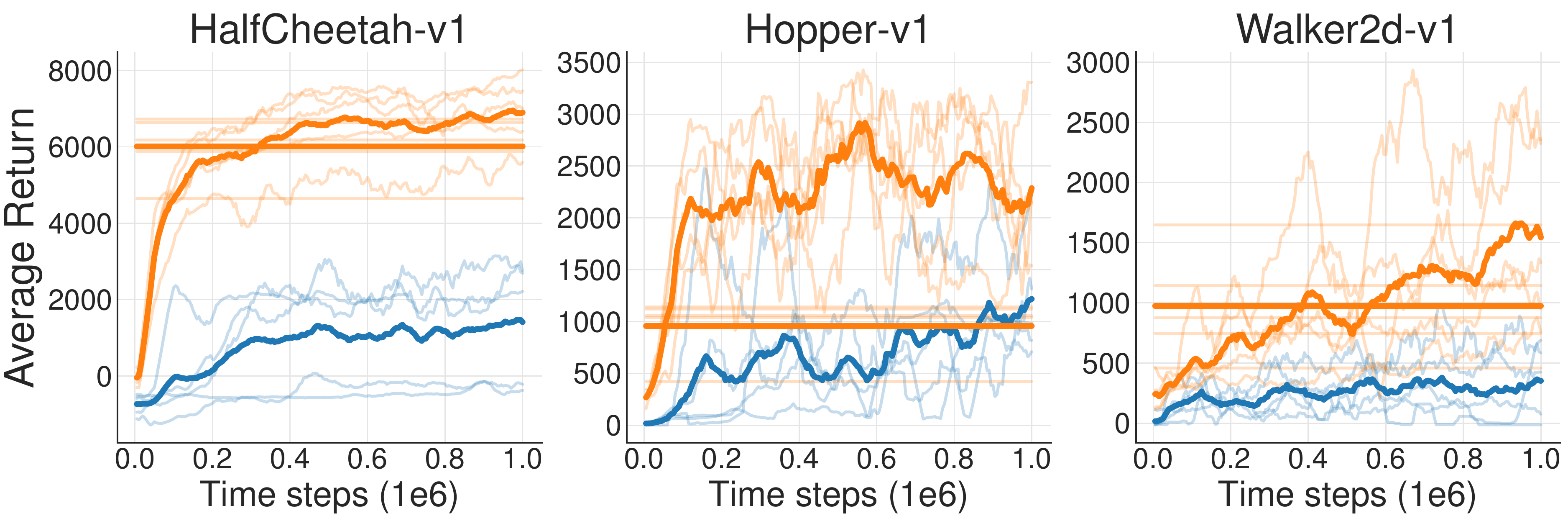}}
\subfloat[Final buffer value estimates]{\includegraphics[width=0.5\linewidth]{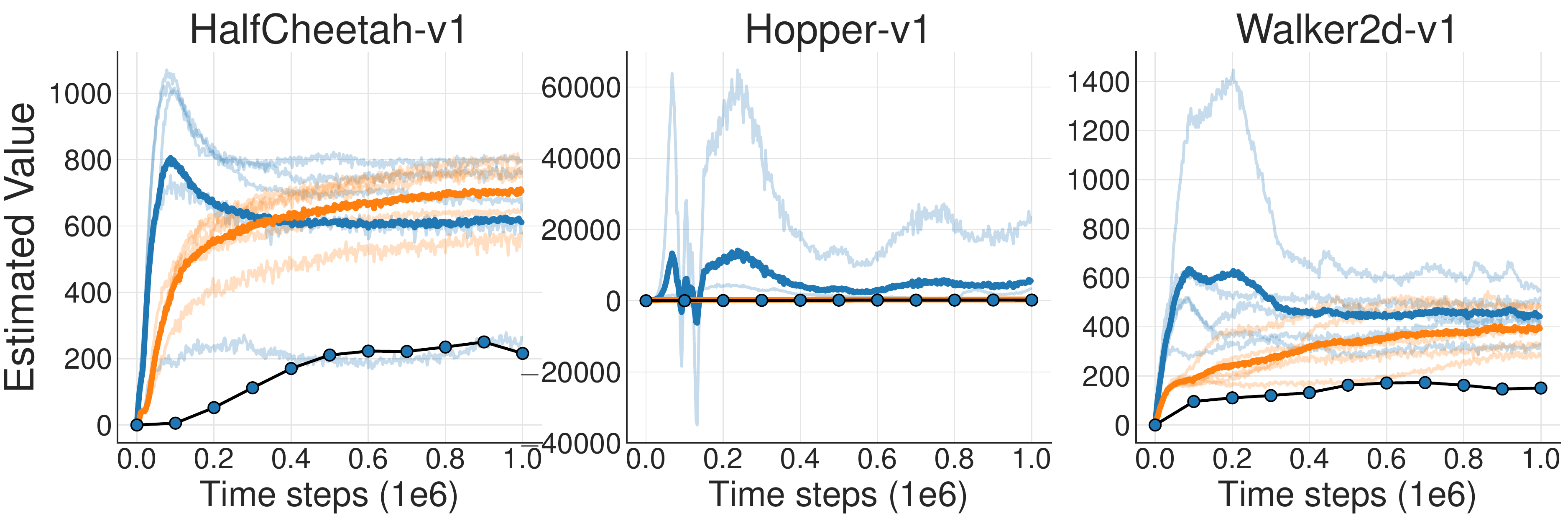}}

\subfloat[Concurrent performance]{\includegraphics[width=0.5\linewidth]{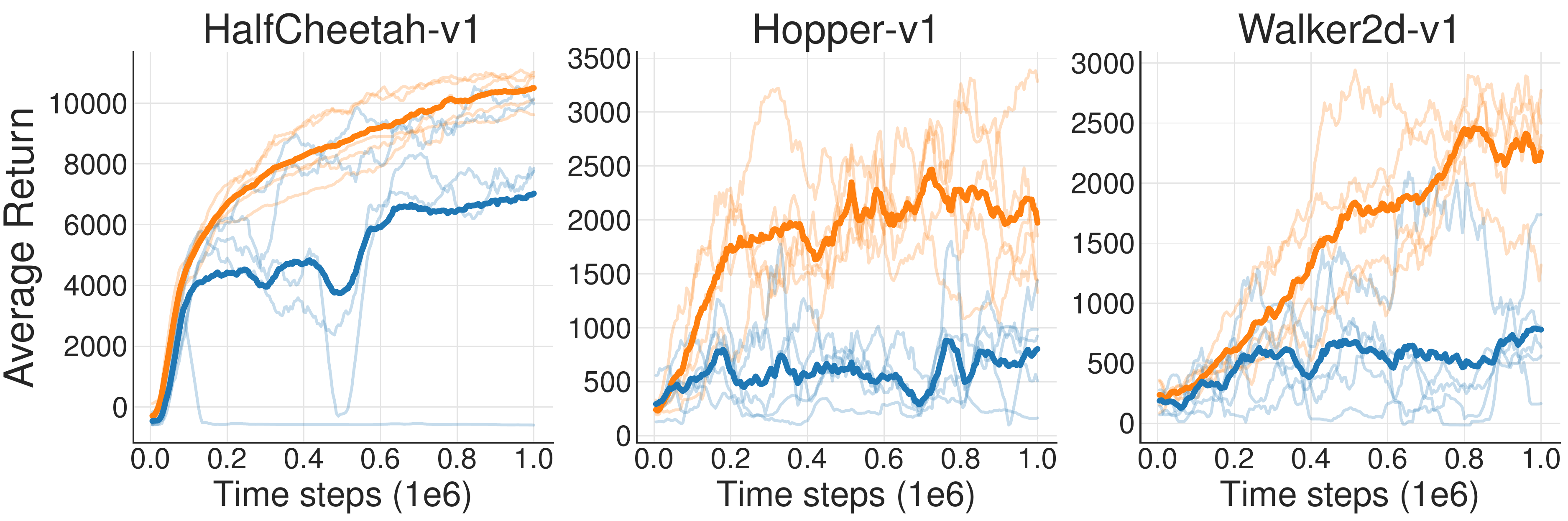}}
\subfloat[Concurrent value estimates]{\includegraphics[width=0.5\linewidth]{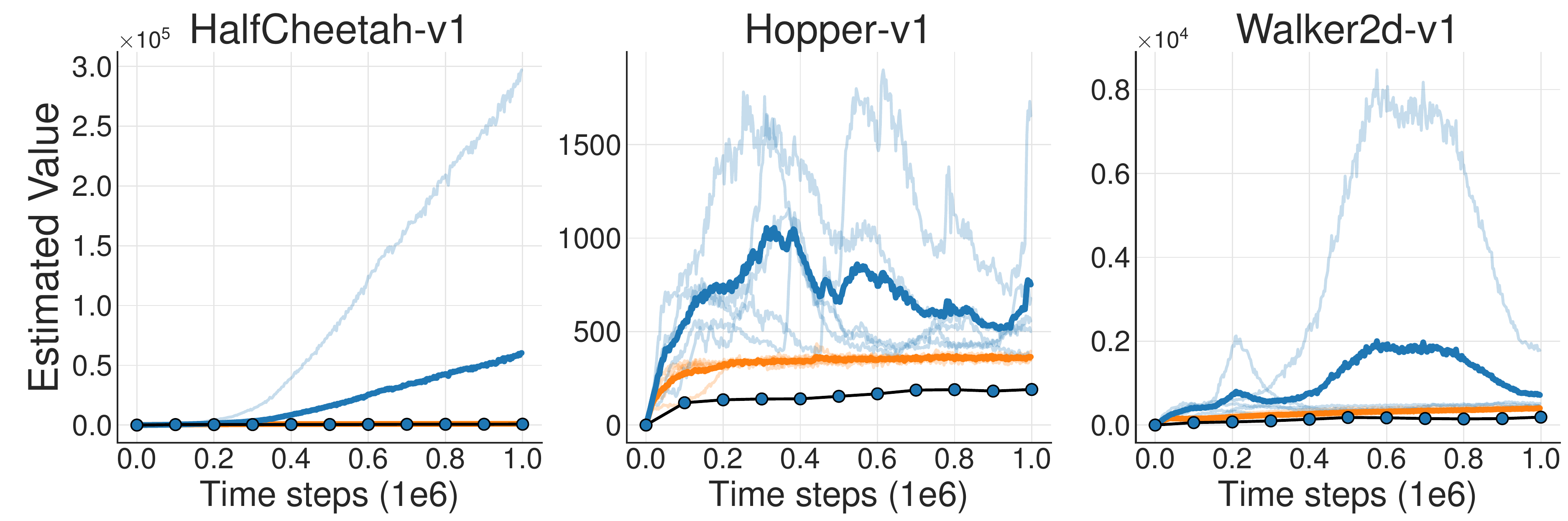}}

\subfloat[Imitation performance]{\includegraphics[width=0.5\linewidth]{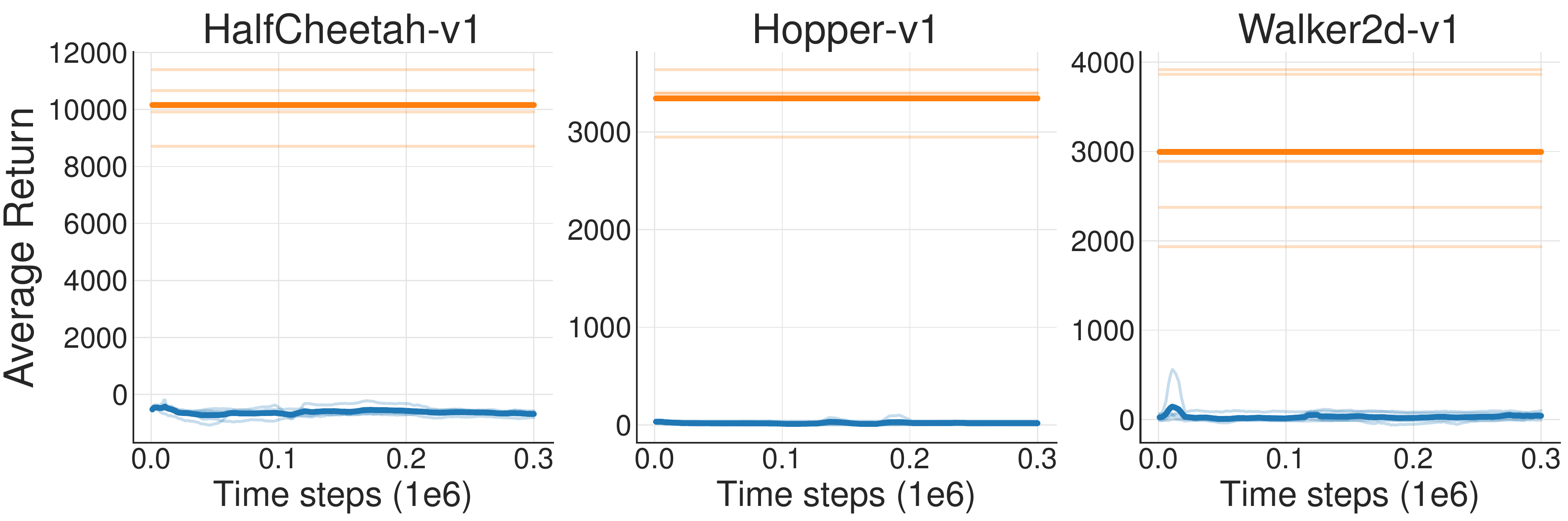}}
\subfloat[Imitation value estimates]{\includegraphics[width=0.5\linewidth]{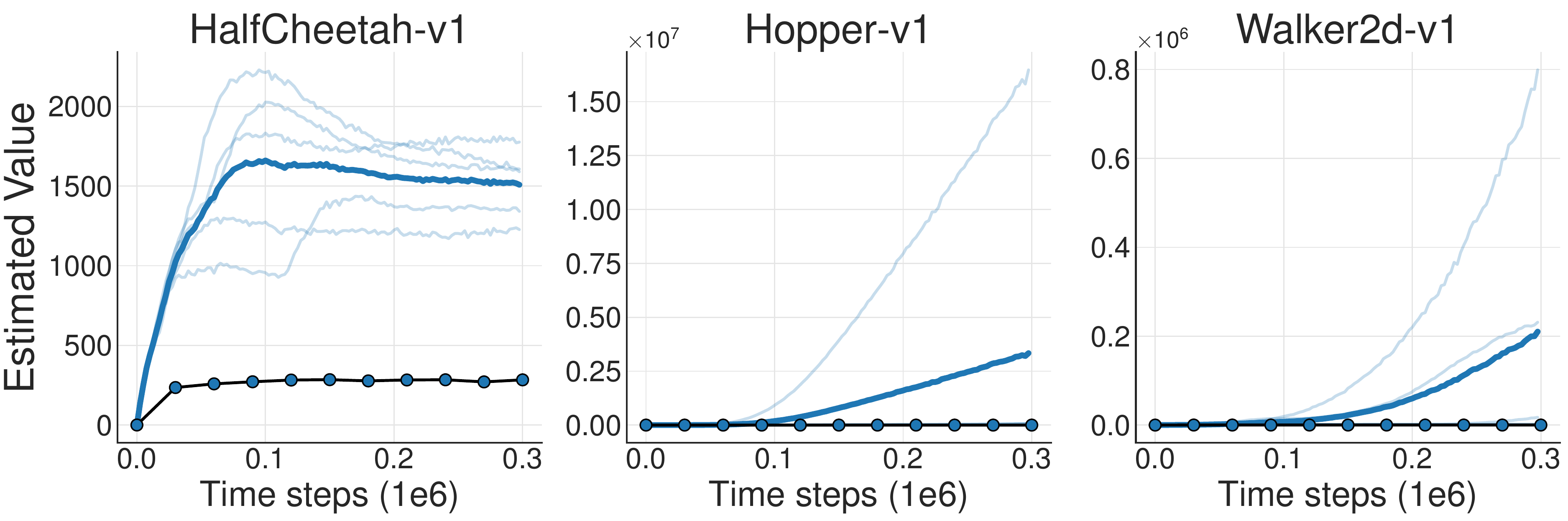}}

\caption{We examine the performance of DDPG in three batch tasks. Each individual trial is plotted with a thin line, with the mean in bold (evaluated without exploration noise). Straight lines represent the average return of episodes contained in the batch (with exploration noise). An estimate of the true value of the off-policy agent, evaluated by Monte Carlo returns, is marked by a dotted line. In the final buffer experiment, the off-policy agent learns from a large, diverse dataset, but exhibits poor learning behavior and value estimation. In the concurrent setting the agent learns alongside a behavioral agent, with access to the same data, but suffers in performance. In the imitation setting, the agent receives data from an expert policy but is unable to learn, and exhibits highly divergent value estimates.}
\end{figure}

\clearpage

\subsection{Complete Experimental Results}

\begin{figure}[ht]
\centering

\subfloat{\includegraphics[width=0.5\linewidth]{results/legend_scores.png}}
\subfloat{\includegraphics[width=0.5\linewidth]{results/legend_val.png}}

\setcounter{subfigure}{0}

\subfloat[Final buffer performance]{\includegraphics[width=0.5\linewidth]{ICML_results/fixed_buffer_score.pdf}}
\subfloat[Final buffer value estimates]{\includegraphics[width=0.5\linewidth]{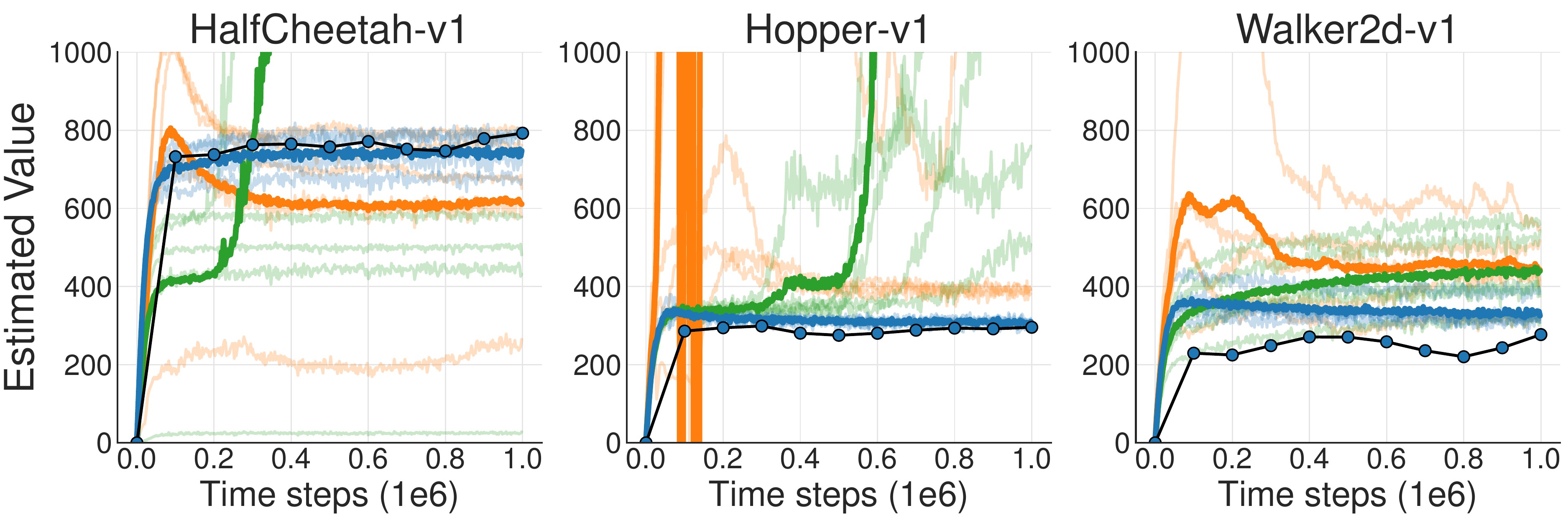}}

\subfloat[Concurrent performance]{\includegraphics[width=0.5\linewidth]{ICML_results/concurrent_score.pdf}}
\subfloat[Concurrent value estimates]{\includegraphics[width=0.5\linewidth]{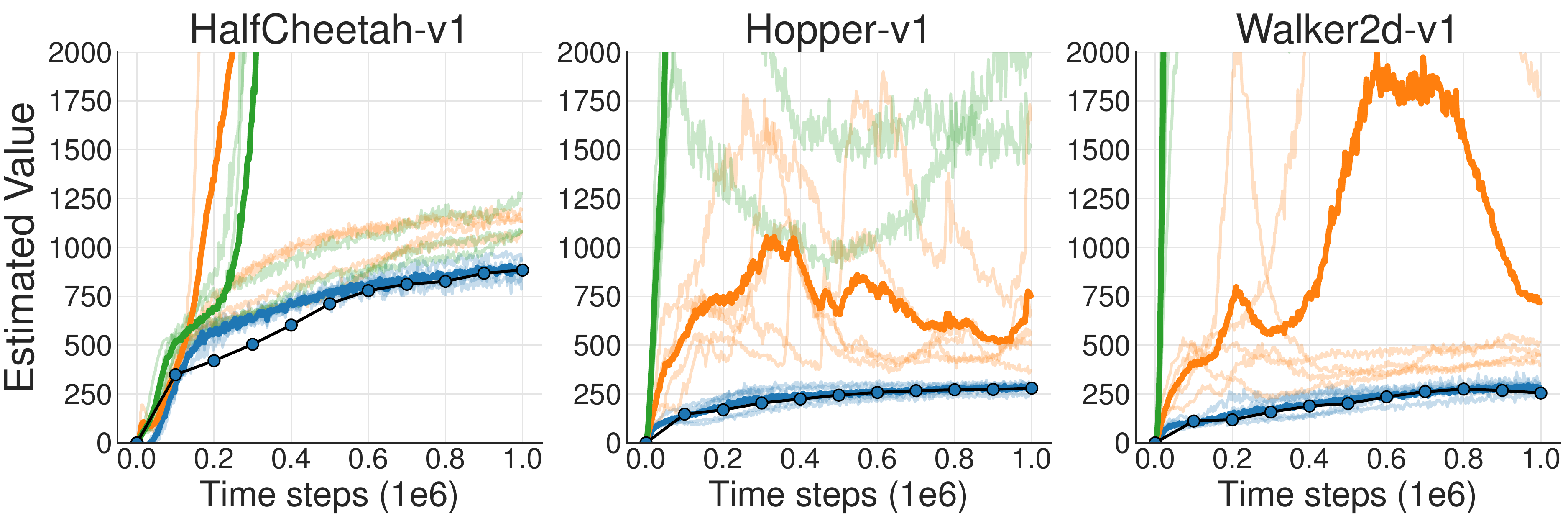}}

\subfloat[Imitation performance]{\includegraphics[width=0.5\linewidth]{ICML_results/imitation_score.pdf}}
\subfloat[Imitation value estimates]{\includegraphics[width=0.5\linewidth]{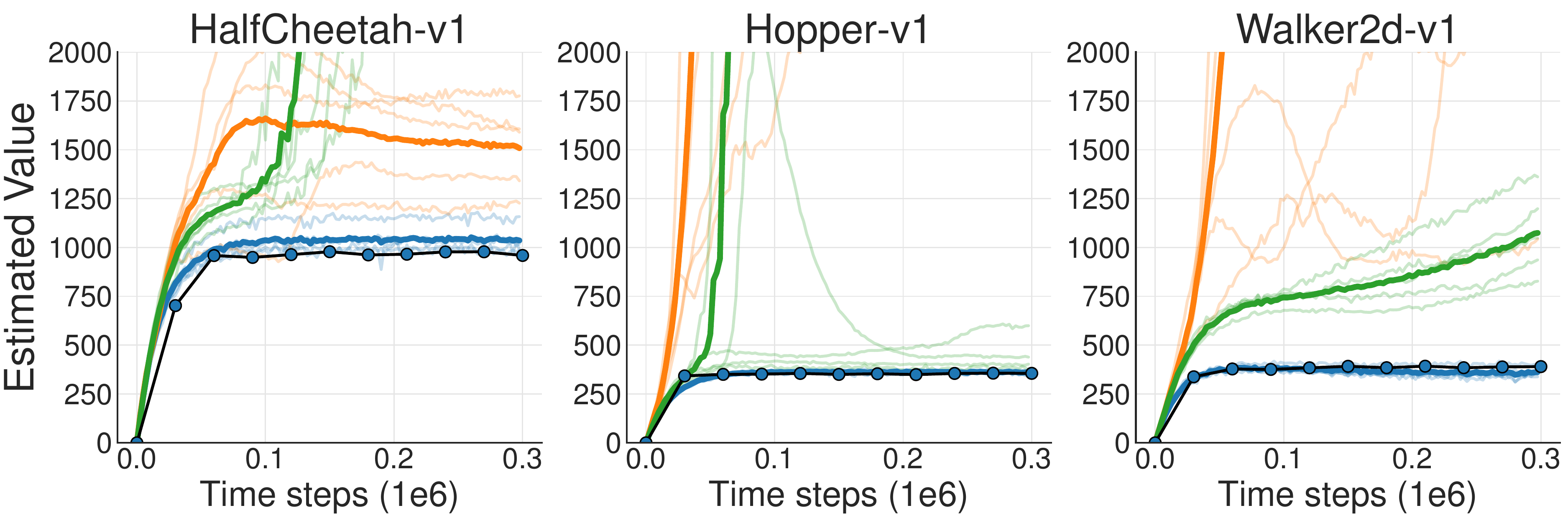}}

\subfloat[Imperfect demonstrations performance]{\includegraphics[width=0.5\linewidth]{ICML_results/robust_score.pdf}}
\subfloat[Imperfect demonstrations value estimates]{\includegraphics[width=0.5\linewidth]{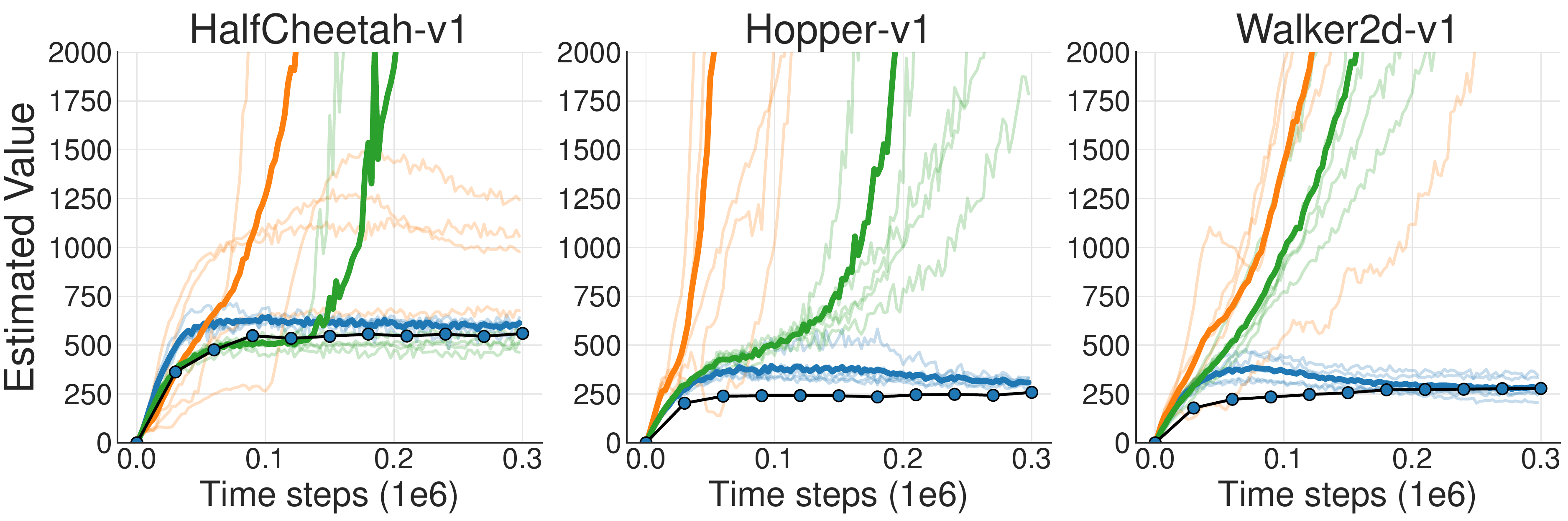}}

\caption{We evaluate BCQ and several baselines on the experiments from Section 3.1, as well as a new imperfect demonstration task. Performance is graphed on the left, and value estimates are graphed on the right. The shaded area represents half a standard deviation. The bold black line measures the average return of episodes contained in the batch. For the value estimates, each individual trial is plotted, with the mean in bold. An estimate of the true value of BCQ, evaluated by Monte Carlo returns, is marked by a dotted line. Only BCQ matches or outperforms the performance of the behavioral policy in all tasks, while exhibiting a highly stable value function in each task.} \label{appendix:complete_results}

\end{figure}

We present the complete set of results across each task and environment in Figure \ref{appendix:complete_results}. These results show that BCQ successful mitigates extrapolation error and learns from a variety of fixed batch settings. Although BCQ with our current hyper-parameters was never found to fail, we noted with slight changes to hyper-parameters, BCQ failed periodically on the concurrent learning task in the HalfCheetah-v1 environment, exhibiting instability in the value function after $750,000$ or more iterations on some seeds. We hypothesize that this instability could occur if the generative model failed to output in-distribution actions, and could be corrected through additional training or improvements to the vanilla VAE. Interestingly, BCQ still performs well in these instances, due to the behavioral cloning-like elements in the algorithm. 

\clearpage

\section{Extrapolation Error in Kernel-Based Reinforcement Learning}

This problem of extrapolation persists in traditional batch reinforcement learning algorithms, such as kernel-based reinforcement learning (KBRL) \citep{ormoneit2002kernel}. For a given batch $\B$ of transitions $(s,a,r,s')$, non-negative density function $\phi: \mathbb{R}^+ \rightarrow \mathbb{R}^+$, hyper-parameter $\tau \in \mathbb{R}$, and norm $|| \cdot ||$, KBRL evaluates the value of a state-action pair (s,a) as follows: 
\begin{align} \label{eqn:KBRL}
Q(s,a) &= \sum_{(s_\B^a, a, r, s_\B') \in \B} \kappa^a_\tau(s,s_\B^a)[r + \y V(s_\B')], \\
\kappa_\tau^a (s,s_\B^a) &= \frac{k_\tau(s,s_\B^a)}{\sum_{\tilde s_\B^a} k_\tau(s, \tilde s_\B^a)}, \qquad k_\tau(s,s_\B^a) = \phi \lp \frac{||s - s_\B^a||}{\tau} \rp,
\end{align}
where $s_\B^a \in \mathcal{S}$ represents states corresponding to the action $a$ for some tuple $(s_\B, a) \in \B$, and $V(s_\B') = \max_{a \text{ s.t.} (s_\B', a) \in \B} Q(s_\B', a)$. At each iteration, KBRL updates the estimates of $Q(s_\B,a_\B)$ for all $(s_\B,a_\B) \in \B$ following Equation (\ref{eqn:KBRL}), then updates $V(s_\B')$ by evaluating $Q(s_\B',a)$ for all $s_\B \in \B$ and $a \in \mathcal{A}$.

\begin{figure}[ht]
\centering
\begin{tikzpicture}[auto,node distance=8mm,>=latex]
  \tikzstyle{round}=[thick,draw=black,circle]

  \node[round,minimum size=1cm] (s0) {$s_0$};
  \node[round,minimum size=1cm,right=25mm of s0] (s1) {$s_1$};

  \draw[->] ([yshift=1ex] s0.east) -- node [above,midway] {$a_1, r=1$} ([yshift=1ex] s1.west);
  \draw[->] ([yshift=-1ex] s1.west) -- node [below,midway] {$a_0, r=0$} ([yshift=-1ex] s0.east);
  \draw[->] (s0) [out=120,in=180,loop] to node [left,midway] {$a_0, r=0$} (s0);
  \draw[->] (s1) [out=60,in=360,loop] to node [right,midway] {$a_1, r=0$} (s1);
\end{tikzpicture}
\caption{Toy MDP with two states $s_0$ and $s_1$, and two actions $a_0$ and $a_1$. Agent receives reward of $1$ for selecting $a_1$ at $s_0$ and $0$ otherwise.}
\label{fig:MDP}
\end{figure}
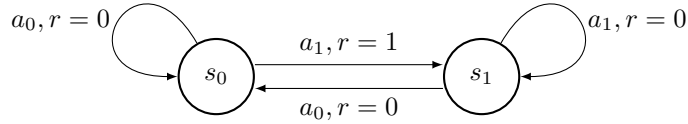

Given access to the entire deterministic MDP, KBRL will provable converge to the optimal value, however when limited to only a subset, we find the value estimation susceptible to extrapolation.  
In Figure \ref{fig:MDP}, we provide a deterministic two state, two action MDP in which KBRL fails to learn the optimal policy when provided with state-action pairs from the optimal policy. Given the batch $\{ (s_0, a_1, r=1, s_1), (s_1, a_0, r=0, s_0) \}$, corresponding to the optimal behavior, and noting that there is only one example of each action, Equation (\ref{eqn:KBRL}) provides the following: 
\begin{equation} \label{eqn:KBRL_updates}
Q(\cdot, a_1) = 1 + \y V(s_1) = 1 + \y Q(s_1,a_0), \qquad  Q(\cdot, a_0) = \y V(s_0) = \y Q(s_0,a_1).
\end{equation}
After sufficient iterations KBRL will converge correctly to $Q(s_0, a_1) = \frac{1}{1 - \y^2}$, $Q(s_1, a_0) = \frac{\y}{1 - \y^2}$. However, when evaluating actions, KBRL erroneously extrapolates the values of each action $Q(\cdot, a_1) = \frac{1}{1 - \y^2}$, $Q(\cdot, a_0) = \frac{\y}{1 - \y^2}$, and its behavior, $\argmax_a Q(s,a)$, will result in the degenerate policy of continually selecting $a_1$. KBRL fails this example by estimating the values of unseen state-action pairs. In methods where the extrapolated estimates can be included into the learning update, such fitted Q-iteration or DQN \citep{ernst2005tree,DQN}, this could cause an unbounded sequence of value estimates, as demonstrated by our results in Section 3.1.


%
 
\clearpage

\section{Additional Experiments} 

\subsection{Ablation Study of Perturbation Model}

\begin{figure}[ht]
\centering

\subfloat{\includegraphics[width=0.65\linewidth]{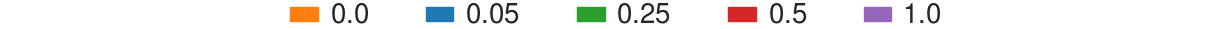}}

\setcounter{subfigure}{0}

\subfloat[Imitation performance]{\includegraphics[width=0.5\linewidth]{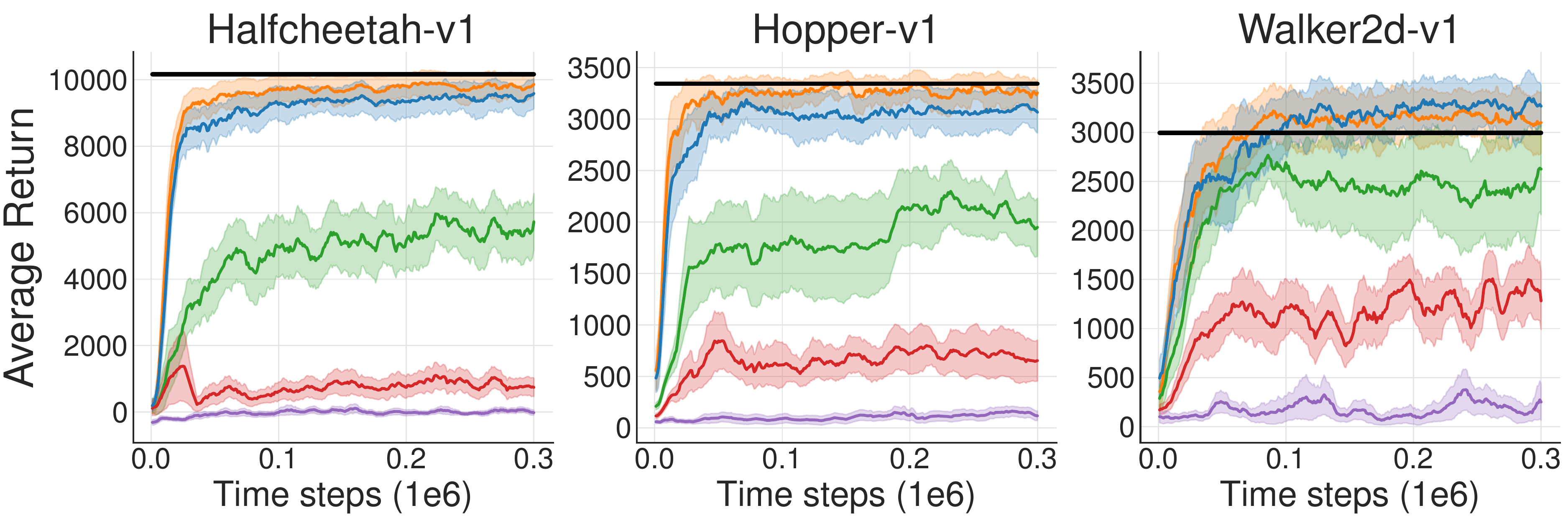}}
\subfloat[Imitation value estimates]{\includegraphics[width=0.5\linewidth]{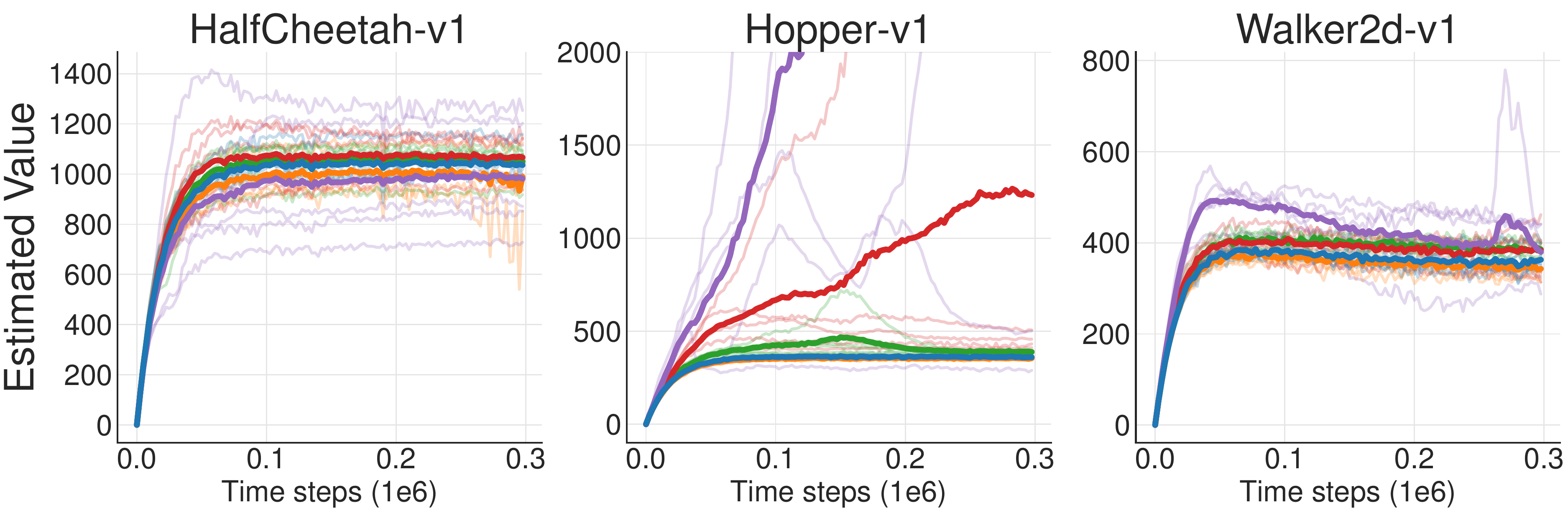}}
\caption{We perform an ablation study on the perturbation model of BCQ, on the imitation task from Section 3.1. Performance is graphed on the left, and value estimates are graphed on the right. The shaded area represents half a standard deviation. The bold black line measures the average return of episodes contained in the batch. For the value estimates, each individual trial is plotted, with the mean in bold.}
\label{fig:ablation}
\end{figure}

BCQ includes a perturbation model $\xi_\ta(s,a, \Phi)$ which outputs a small residual update to the actions sampled by the generative model in the range $[-\Phi, \Phi]$. This enables the policy to select actions which may not have been sampled by the generative model. If $\Phi = a_{\text{max}} - a_{\text{min}}$, then all actions can be plausibly selected by the model, similar to standard deep reinforcement learning algorithms, such as DQN and DDPG \cite{DQN, DDPG}. In Figure \ref{fig:ablation} we examine the performance and value estimates of BCQ when varying the hyper-parameter $\Phi$, which corresponds to how much the model is able to move away from the actions sampled by the generative model. 

We observe a clear drop in performance with the increase of $\Phi$, along with an increase in instability in the value function. Given that the data is based on expert performance, this is consistent with our understanding of extrapolation error. With larger $\Phi$ the agent learns to take actions that are further away from the data in the batch after erroneously overestimating the value of suboptimal actions. This suggests the ideal value of $\Phi$ should be small enough to stay close to the generated actions, but large enough such that learning can be performed when exploratory actions are included in the dataset. 

\newpage

\subsection{Uncertainty Estimation for Batch-Constrained Reinforcement Learning} 

Section 4.2 proposes using generation as a method for constraining the output of the policy $\pi$ to eliminate actions which are unlikely under the batch. However, a more natural approach would be through approximate uncertainty-based methods \cite{osband2016deep, gal2016improving, azizzadenesheli2018bayesian}. These methods are well-known to be effective for \textit{exploration}, however we examine their properties for the \textit{exploitation} where we would like to avoid uncertain actions. 

To measure the uncertainty of the value network, we use ensemble-based methods, with an ensemble of size $4$ and $10$ to mimic the models used by \citet{buckman2018sample} and \citet{osband2016deep} respectively. Each network is trained with separate mini-batches, following the standard deep Q-learning update with a target network. These networks use the default architecture and hyper-parameter choices as defined in Section \ref{appendix:implementation}. The policy $\pi_\phi$ is trained to minimize the standard deviation $\sigma$ across the ensemble:
\begin{equation}
    \phi \leftarrow \argmin_\phi \sum_{(s,a) \in \B} \sigma \lp \{ Q_{\ta_i}(s,a) \}_{i=1}^N \rp .
\end{equation}
If the ensembles were a perfect estimate of the uncertainty, the policy would learn to select the most certain action for a given state, minimizing the extrapolation error and effectively imitating the data in the batch. 

\begin{figure}[ht]
\centering
\includegraphics[trim={75mm 0 0 0}, clip, width=0.5\linewidth ]{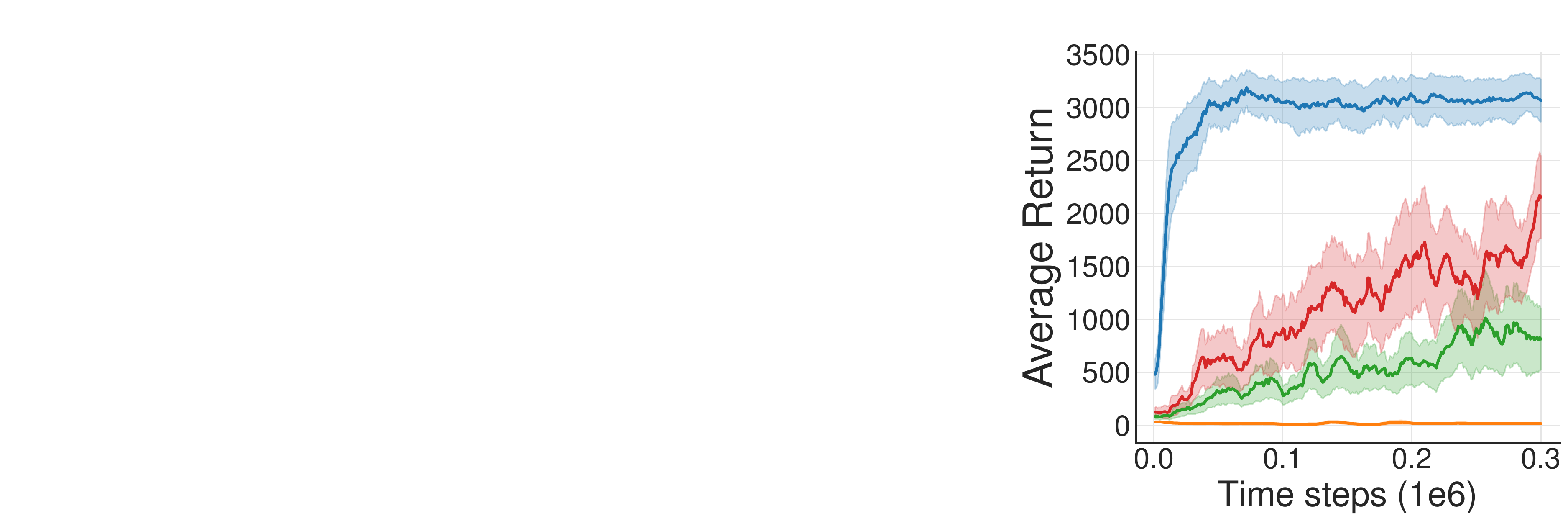}\includegraphics[trim={0 0 75mm 0}, clip, width=0.5\linewidth]{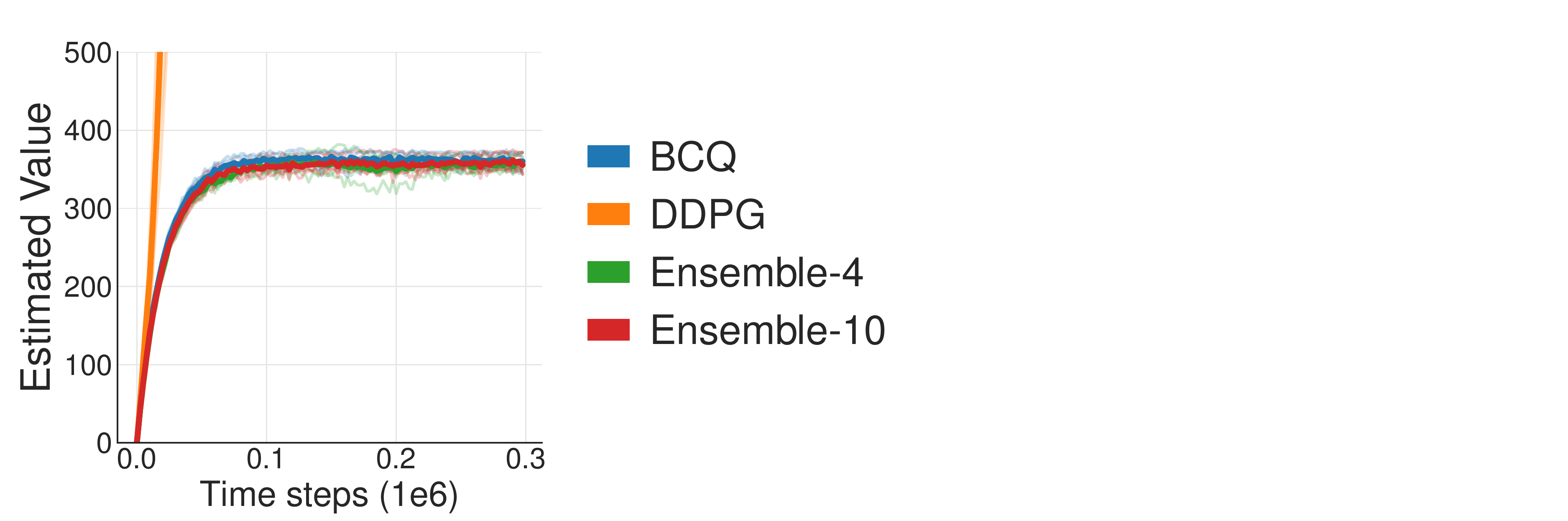}
\vspace{-8mm}

\hspace{0.26\linewidth}\subfloat[Imitation performance]{\hspace{0.30\linewidth}}\hspace{-0.27\linewidth}\subfloat[Imitation value estimates]{\hspace{0.71\linewidth}}
\caption{A comparison with uncertainty-based methods on the Hopper-v1 environment from OpenAI gym, on the imitation task. Although ensemble based methods are able to generate stable value functions (right) under these conditions, they fail to constrain the action space to the demonstrated expert actions and suffer in performance compared to our approach, BCQ (left).}
\label{fig:uncertainty}
\end{figure}

To test these uncertainty-based methods, we examine their performance on the \textit{imitation} task in the Hopper-v1 environment \cite{mujoco, OpenAIGym}. In which a dataset of 1 million expert transitions are provided to the agents. Additional experimental details can be found in Section \ref{appendix:experimental}. The performance, alongside the value estimates of the agents are displayed in Figure \ref{fig:uncertainty}.  


We find that neither ensemble method is sufficient to constrain the action space to only the expert actions. However, the value function is stabilized, suggesting that ensembles are an effective strategy for eliminating outliers or large deviations in the value from erroneous extrapolation. Unsurprisingly, the large ensemble provides a more accurate estimate of the uncertainty. While scaling the size of the ensemble to larger values could possibly enable an effective batch-constraint, increasing the size of the ensemble induces a large computational cost. Finally, in this task, where only expert data is provided, the policy can attempt to imitate the data without consideration of the value, however in other tasks, a weighting between value and uncertainty would need to be carefully tuned. On the other hand, BCQ offers a computational cheap approach without requirements for difficult hyper-parameter tuning. 

\clearpage

\subsection{Random Behavioral Policy Study} \label{appendix:random}

The experiments in Section 3.1 and 5 use a learned, or partially learned, behavioral policy for data collection. This is a necessary requirement for learning meaningful behavior, as a random policy generally fails to provide sufficient coverage over the state space. However, in simple toy problems, such as the pendulum swing-up task and the reaching task with a two-joint arm from OpenAI gym \citep{OpenAIGym}, a random policy can sufficiently explore the environment, enabling us to examine the properties of algorithms with entirely non-expert data.

\begin{figure}[ht]
\centering
\includegraphics[width=0.5\linewidth]{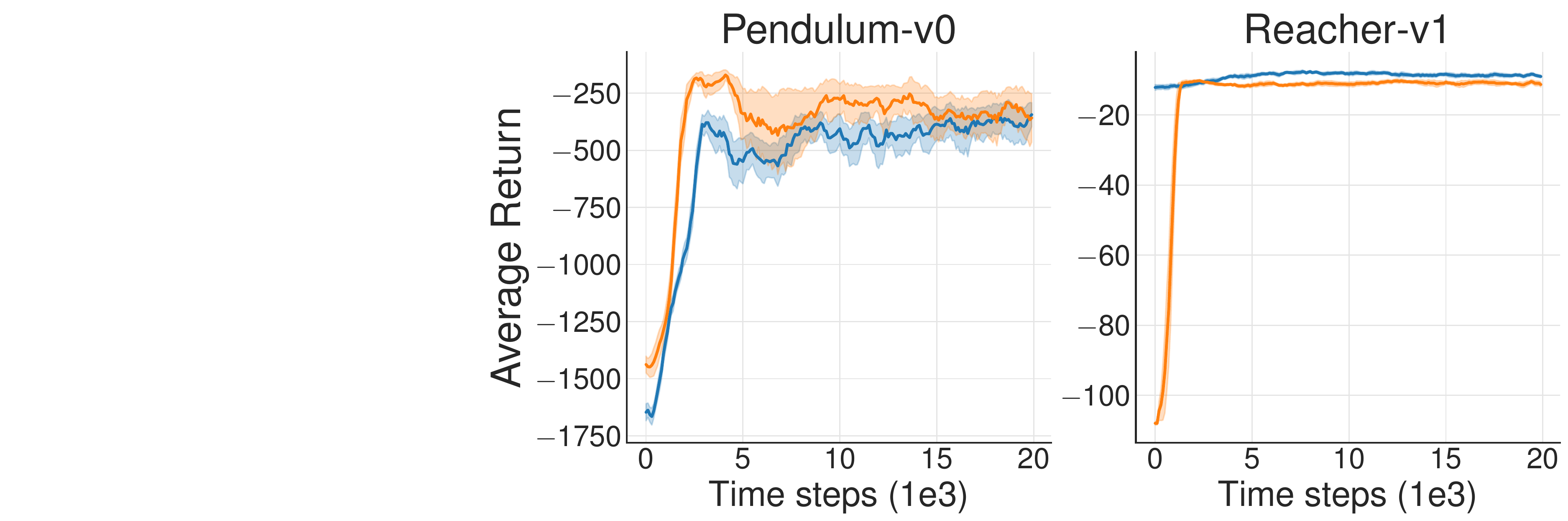}\includegraphics[width=0.5\linewidth]{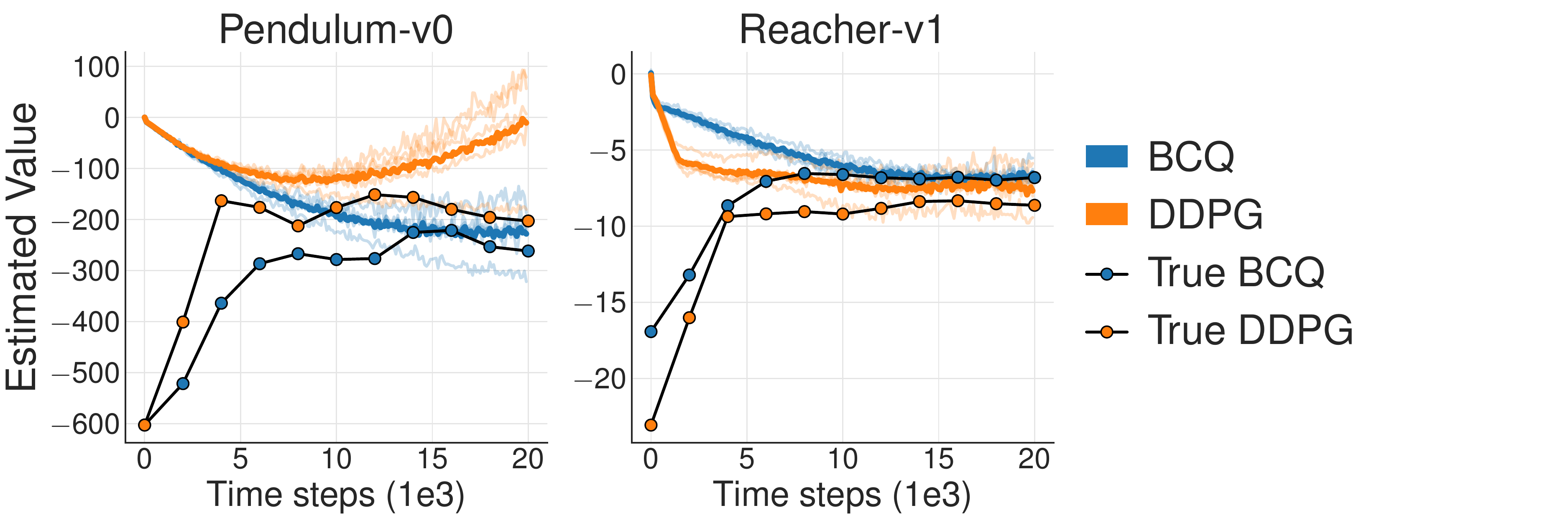}
\vspace{-8mm}

\hspace{0.20\linewidth}\subfloat[Random behavioral performance]{\hspace{0.30\linewidth}}\hspace{-0.13\linewidth}\subfloat[Random behavioral value estimates]{\hspace{0.63\linewidth}}
\caption{We evaluate BCQ and DDPG on a batch collected by a random behavioral policy. The shaded area represents half a standard deviation. Value estimates include a plot of each trial, with the mean in bold. An estimate of the true value of BCQ and DDPG, evaluated by Monte Carlo returns, is marked by a dotted line. While both BCQ and DDPG perform well, BCQ exhibits more stability in the value function for the Pendulum task, and outperforms DDPG in the Reacher task. These tasks demonstrate examples where any imitation learning would fail as the data collection procedure is entirely random.}
\label{fig:random}
\end{figure}

In Figure \ref{fig:random}, we examine the performance of our algorithm, BCQ, as well as DDPG \citep{DDPG}, on these two toy problems, when learning off-policy from a small batch of 5000 time steps, collected entirely by a random policy. We find that both BCQ and DDPG are able to learn successfully in these off-policy tasks. These results suggest that BCQ is less restrictive than imitation learning algorithms, which require expert data to learn. We also find that unlike previous environments, given the small scale of the state and action space, the random policy is able to provide sufficient coverage for DDPG to learn successfully. 


\section{Missing Background} 

\subsection{Variational Auto-Encoder} \label{appendix:VAE}

A variational auto-encoder (VAE) \citep{kingma2013auto} is a generative model which aims to maximize the marginal log-likelihood $\log p(X) = \sum_{i=1}^N \log p(x_i)$ where $X = \{x_1, ..., x_N\}$, the dataset. While computing the marginal likelihood is intractable in nature, we can instead optimize the variational lower-bound: 
\begin{equation} \label{vae}
\log p(X) \geq \E_{q(X|z)}[ \log p(X|z) ] + D_{\text{KL}}(q(z|X)||p(z)),
\end{equation}
where $p(z)$ is chosen a prior, generally the multivariate normal distribution $\N(0,I)$. We define the posterior $q(z|X) = \N(z|\mu(X), \s^2(X)I)$ as the encoder and $p(X|z)$ as the decoder. Simply put, this means a VAE is an auto-encoder, where a given sample $x$ is passed through the encoder to produce a random latent vector $z$, which is given to the decoder to reconstruct the original sample $x$. The VAE is trained on a reconstruction loss, and a KL-divergence term according to the distribution of the latent vectors. To perform gradient descent on the variational lower bound we can use the re-parametrization trick \citep{kingma2013auto,rezende2014stochastic}:
\begin{equation}
\E_{z \sim \N(\mu, \s)}[f(z)] = \E_{\e \sim \N(0, I)}[f(\mu + \s \e)].
\end{equation}
This formulation allows for back-propagation through stochastic nodes, by noting $\mu$ and $\s$ can be represented by deterministic functions. During inference, random values of $z$ are sampled from the multivariate normal and passed through the decoder to produce samples $x$. 

\clearpage

\section{Experimental Details} \label{appendix:experimental}

Each environment is run for 1 million time steps, unless stated otherwise, with evaluations every 5000 time steps, where an evaluation measures the average reward from 10 episodes with no exploration noise. Our results are reported over 5 random seeds of the behavioral policy, OpenAI Gym simulator and network initialization. Value estimates are averaged over mini-batches of $100$ and sampled every $2500$ iterations. The \textit{true} value is estimated by sampling $100$ state-action pairs from the buffer replay and computing the discounted return by running the episode until completion while following the current policy.

Each agent is trained after each episode by applying one training iteration per each time step in the episode. The agent is trained with transition tuples $(s,a,r,s')$ sampled from an experience replay that is defined by each experiment. We define four possible experiments. Unless stated otherwise, default implementation settings, as defined in Section \ref{appendix:implementation}, are used.

\textbf{Batch 1 (Final buffer).} We train a DDPG \citep{DDPG} agent for 1 million time steps, adding large amounts of Gaussian noise $(\N(0,0.5))$ to induce exploration, and store all experienced transitions in a buffer replay. This training procedure creates a buffer replay with a diverse set of states and actions. A second, randomly initialized agent is trained using the 1 million stored transitions. 

\textbf{Batch 2 (Concurrent learning).} We simultaneously train two agents for 1 million time steps, the first DDPG agent, performs data collection and each transition is stored in a buffer replay which both agents learn from. This means the behavioral agent learns from the standard training regime for most off-policy deep reinforcement learning algorithms, and the second agent is learning off-policy, as the data is collected without direct relationship to its current policy. Batch 2 differs from Batch 1 as the agents are trained with the same version of the buffer, while in Batch 1 the agent learns from the final buffer replay after the behavioral agent has finished training.

\textbf{Batch 3 (Imitation).} A DDPG agent is trained for 1 million time steps. The trained agent then acts as an expert policy, and is used to collect a dataset of 1 million transitions. This dataset is used to train a second, randomly initialized agent. In particular, we train DDPG across 15 seeds, and select the 5 top performing seeds as the expert policies. 

\textbf{Batch 4 (Imperfect demonstrations). } The expert policies from Batch 3 are used to collect a dataset of 100k transitions, while selecting actions randomly with probability $0.3$ and adding Gaussian noise $\N(0,0.3)$ to the remaining actions. This dataset is used to train a second, randomly initialized agent. 

\section{Implementation Details} \label{appendix:implementation}

Across all methods and experiments, for fair comparison, each network generally uses the same hyper-parameters and architecture, which are defined in Table~\ref{table:default_hp} and Figure~\ref{fig:default_arch} respectively. The value functions follow the standard practice \citep{DQN} in which the Bellman update differs for terminal transitions. When the episode ends by reaching some terminal state, the value is set to $0$ in the learning target $y$: 
\begin{equation}
y = 
  \begin{cases}
  r  & \text{if terminal } s' \\
  r + \y Q_{\theta'}(s',a') & \text{else} \\
  \end{cases}
\end{equation}
Where the termination signal from time-limited environments is ignored, thus we only consider a state $s_t$ terminal if $t < $ \texttt{max horizon}. 

\begin{table}[ht]
\centering
\caption{Default hyper-parameters.} \label{table:default_hp}
\begin{center}
\begin{small}
\begin{tabular}{lc}
\toprule
Hyper-parameter & Value \\
\midrule
Optimizer & Adam \cite{adam} \\
Learning Rate & $10^{-3}$ \\
Batch Size & $100$ \\
Normalized Observations & False \\
Gradient Clipping & False \\
Discount Factor & $0.99$ \\
Target Update Rate ($\tau$) & $0.005$ \\ 
Exploration Policy & $\N(0, 0.1)$ \\ 
\bottomrule
\end{tabular}
\end{small}
\end{center}
\end{table}

\begin{figure}[ht]
\centering
\begin{BVerbatim}
(input dimension, 400)
ReLU
(400, 300)
RelU
(300, output dimension)
\end{BVerbatim}
\caption{Default network architecture, based on DDPG \cite{DDPG}. All actor networks are followed by a \texttt{tanh} $\cdot$ \texttt{max action size}} \label{fig:default_arch}
\end{figure}

\textbf{BCQ.} BCQ uses four main networks: a perturbation model $\xi_\phi(s,a)$, a state-conditioned VAE $G_\w(s)$ and a pair of value networks $Q_{\ta_1}(s,a), Q_{\ta_2}(s,a)$. Along with three corresponding target networks $\xi_{\phi'}(s,a), Q_{\ta'_1}(s,a), Q_{\ta'_2}(s,a)$. Each network, other than the VAE follows the default architecture (Figure \ref{fig:default_arch}) and the default hyper-parameters (Table \ref{table:default_hp}). For $\xi_\phi(s,a, \Phi)$, the constraint $\Phi$ is implemented through a tanh activation multiplied by $I \cdot \Phi$ following the final layer. 

As suggested by \citet{fujimoto2018addressing}, the perturbation model is trained only with respect to $Q_{\ta_1}$, following the deterministic policy gradient algorithm \citep{DPG}, by performing a residual update on a single action sampled from the generative model:
\begin{equation}
\begin{split}
\phi &\leftarrow \argmax_{\phi} \sum_{(s,a) \in \B} Q_{\ta_1}(s,a + \xi_{\phi}(s, a, \Phi)), \\
&a \sim G_\w(s).
\end{split}
\end{equation}

To penalize uncertainty over future states, we train a pair of value estimates $\{ Q_{\ta_1}, Q_{\ta_2} \}$ and take a weighted minimum between the two values as a learning target $y$ for both networks. First $n$ actions are sampled with respect to the generative model, and then adjusted by the target perturbation model, before passed to each target Q-network:
\begin{equation} \label{appendix:softmin}
\begin{split}
y &= r + \y \max_{ai} \lb \lambda \min_{j=1,2} Q_{\ta'_j}(s', \tilde a_i) + (1 - \lambda) \max_{j=1,2} Q_{\ta'_j}(s', \tilde a_i) \rb, \\ 
& \{ \tilde a_i = a_i + \xi_{\phi'}(s, a_i, \Phi) \}_{i=0}^n, \qquad \{ a_i \sim G_\w(s') \}_{i=0}^n, \\
&\mathcal{L}_\text{value,i} = \sum_{(s,a,r,s') \in \B} (y - Q_{\ta_i}(s,a))^2.
\end{split}
\end{equation}
Both networks are trained with the same target $y$, where $\lambda=0.75$. 

The VAE $G_\w$ is defined by two networks, an encoder $E_{\w_1}(s,a)$ and decoder $D_{\w_2}(s, z)$, where $\w = \{\w_1, \w_2\}$. The encoder takes a state-action pair and outputs the mean $\mu$ and standard deviation $\sigma$ of a Gaussian distribution $\N(\mu, \sigma)$. The state $s$, along with a latent vector $z$ is sampled from the Gaussian, is passed to the decoder $D_{\w_2}(s, z)$ which outputs an action. Each network follows the default architecture (Figure \ref{fig:default_arch}), with two hidden layers of size $750$, rather than $400$ and $300$. The VAE is trained with respect to the mean squared error of the reconstruction along with a KL regularization term:
\begin{align}
\mathcal{L}_{\text{reconstruction}} &= \sum_{(s,a) \in \B} (D_{\w_2}(s,z) - a)^2, \quad z = \mu + \sigma \cdot \e, \quad \e \sim \N(0,1), \\
\mathcal{L}_{\text{KL}} &= D_{\text{KL}}(\N(\mu, \sigma)||\N(0,1)), \\
\mathcal{L}_{\text{VAE}} &= \mathcal{L}_{\text{reconstruction}} + \lambda \mathcal{L}_{\text{KL}}.
\end{align}
Noting the Gaussian form of both distributions, the KL divergence term can be simplified \citep{kingma2013auto}:
\begin{equation}
D_{\text{KL}}(\N(\mu, \sigma)||\N(0,1)) = - \frac{1}{2} \sum_{j=1}^J (1 +\log(\sigma_j^2) - \mu_j^2 - \sigma_j^2),
\end{equation}
where $J$ denotes the dimensionality of $z$. For each experiment, $J$ is set to twice the dimensionality of the action space. The KL divergence term in  $\mathcal{L}_{\text{VAE}}$ is normalized across experiments by setting $\lambda = \frac{1}{2J}$. During inference with the VAE, the latent vector $z$ is clipped to a range of $[-0.5, 0.5]$ to limit generalization beyond previously seen actions. 
For the small scale experiments in Supplementary Material \ref{appendix:random}, $L_2$ regularization with weight $10^{-3}$ was used for the VAE to compensate for the small number of samples. The other networks remain unchanged. 
In the value network update (Equation~\ref{appendix:softmin}), the VAE is sampled multiple times and passed to both the perturbation model and each Q-network. For each state in the mini-batch the VAE is sampled $n=10$ times. This can be implemented efficiently by passing a latent vector with batch size $10$ $\cdot$ \texttt{batch size}, effectively $1000$, to the VAE and treating the output as a new mini-batch for the perturbation model and each Q-network. When running the agent in the environment, we sample from the VAE $10$ times, perturb each action with $\xi_\phi$ and sample the highest valued action with respect to $Q_{\ta_1}$. 

\textbf{DDPG.} Our DDPG implementation deviates from some of the default architecture and hyper-parameters to mimic the original implementation more closely \citep{DDPG}. In particular, the action is only passed to the critic at the second layer (Figure \ref{fig:DDPG_arch}), the critic uses $L_2$ regularization with weight $10^{-2}$, and the actor uses a reduced learning rate of $10^{-4}$. 
\begin{figure}[ht]
\centering
\begin{BVerbatim}
(state dimension, 400)
ReLU
(400 + action dimension, 300)
RelU
(300, 1)
\end{BVerbatim}
\caption{DDPG Critic Network Architecture.} \label{fig:DDPG_arch}
\end{figure}

As done by \citet{fujimoto2018addressing}, our DDPG agent randomly selects actions for the first 10k time steps for HalfCheetah-v1, and 1k time steps for Hopper-v1 and Walker2d-v1. This was found to improve performance and reduce the likelihood of local minima in the policy during early iterations.

\textbf{DQN.} Given the high dimensional nature of the action space of the experimental environments, our DQN implementation selects actions over an independently discretized action space. Each action dimension is discretized separately into $10$ possible actions, giving $10J$ possible actions, where $J$ is the dimensionality of the action-space. A given state-action pair $(s,a)$ then corresponds to a set of state-sub-action pairs $(s,a_{ij})$, where $i \in \{1,...,10\}$ bins and $j=\{1,...,J\}$ dimensions. In each DQN update, all state-sub-action pairs $(s,a_{ij})$ are updated with respect to the average value of the target state-sub-action pairs $(s', a'_{ij})$. The learning update of the discretized DQN is as follows:
\begin{align} 
y &= r + \frac{\y}{n} \sum_{j=1}^J \max_{i} Q_{\ta'}(s',a_{ij}'), \\
\ta &\leftarrow \argmin_\ta \sum_{(s,a,r,s') \in \B} \sum_{j=1}^J \lp y - Q_\ta(s,a_{ij}) \rp^2,
\end{align}
Where $a_{ij}$ is chosen by selecting the corresponding bin $i$ in the discretized action space for each dimension $j$. For clarity, we provide the exact DQN network architecture in Figure \ref{fig:DQN_arch}.
\begin{figure}[ht]
\centering
\begin{BVerbatim}
(state dimension, 400)
ReLU
(400, 300)
RelU
(300, 10 action dimension)
\end{BVerbatim}
\caption{DQN Network Architecture.} \label{fig:DQN_arch}
\end{figure}

\textbf{Behavioral Cloning.} We use two behavioral cloning methods, VAE-BC and BC. VAE-BC is implemented and trained exactly as $G_\w(s)$ defined for BCQ. BC uses a feed-forward network with the default architecture and hyper-parameters, and trained with a mean-squared error reconstruction loss. 



\end{document}